%% file: root.tex
\documentclass[shortAfour,sageh,times]{sagej}

\input{macros.tex}

\newcommand\BibTeX{{\rmfamily B\kern-.05em \textsc{i\kern-.025em b}\kern-.08em
T\kern-.1667em\lower.7ex\hbox{E}\kern-.125emX}}

\def\volumeyear{2024}

\begin{document}

\runninghead{Ji et al.}

\title{An Expert Ensemble for Detecting Anomalous Scenes, Interactions, and Behaviors in Autonomous Driving}

\author{Tianchen Ji$^\star$, Neeloy Chakraborty$^\star$, Andre Schreiber, and Katherine Driggs-Campbell}

\affiliation{Coordinated Science Laboratory, University of Illinois at Urbana-Champaign, Urbana, IL, USA \\ \\
$^\star$Tianchen Ji and Neeloy Chakraborty contributed equally.}

\corrauth{Tianchen Ji, Coordinated Science Laboratory,
University of Illinois at Urbana-Champaign,
1308 W Main Street,
Urbana, IL 61801, USA.}

\email{tj12@illinois.edu}

\begin{abstract}
As automated vehicles enter public roads, safety in a near-infinite number of driving scenarios becomes one of the major concerns for the widespread adoption of fully autonomous driving. The ability to detect anomalous situations outside of the operational design domain is a key component in self-driving cars, enabling us to mitigate the impact of abnormal ego behaviors and to realize trustworthy driving systems. On-road anomaly detection in egocentric videos remains a challenging problem due to the difficulties introduced by complex and interactive scenarios. We conduct a holistic analysis of common on-road anomaly patterns, from which we propose three unsupervised anomaly detection experts: a scene expert that focuses on frame-level appearances to detect abnormal scenes and unexpected scene motions; an interaction expert that models normal relative motions between two road participants and raises alarms whenever anomalous interactions emerge; and a behavior expert which monitors abnormal behaviors of individual objects by future trajectory prediction. To combine the strengths of all the modules, we propose an expert ensemble (Xen) using a Kalman filter, in which the final anomaly score is absorbed as one of the states and the observations are generated by the experts. Our experiments employ a novel evaluation protocol for realistic model performance, demonstrate superior anomaly detection performance than previous methods, and show that our framework has potential in classifying anomaly types using unsupervised learning on a large-scale on-road anomaly dataset.
\end{abstract}


\maketitle

\input{Sections/01-Introduction}
\input{Sections/02-RelatedWork}
\input{Sections/03-Overview}
\input{Sections/04-Experts}
\input{Sections/05-Ensemble}
\input{Sections/06-Experiments}
\input{Sections/07-Discussion}
\input{Sections/08-Conclusion}


\begin{dci}
The authors declared no potential conflicts of interest with respect to the research, authorship, and/or publication of this article.
\end{dci}

\begin{funding}
The author(s) disclosed receipt of the following financial support for the research, authorship, and/or publication of this article: This work was supported in part by the National Robotics Initiative 2.0 through USDA/NIFA [\#2021-67021-33449] and the National Science Foundation [\#2143435].
\end{funding}

\theendnotes
\bibliographystyle{SageH}
\bibliography{updatedBibFile}

\end{document}

%% file: macros.tex
\usepackage{moreverb,url}
\usepackage[colorlinks,bookmarksopen,bookmarksnumbered,linkcolor=blue,citecolor=blue,urlcolor=blue]{hyperref}

\usepackage{makecell}
\usepackage{multirow}
\usepackage{mathtools}
\usepackage{subcaption}
\usepackage[inline]{enumitem}

%% file: Sections/01-Introduction.tex
\section{Introduction}
Autonomous driving is at a critical stage in revolutionizing transportation systems and reshaping societal norms. More than 1,400 self-driving cars, trucks, and other vehicles are currently in operation or testing in the U.S.~\citep{etherington2019over}, and 4.5 million autonomous vehicles are expected to run on U.S. roads by 2030~\citep{meyer2023safety}. While autonomous driving is promising in improving traffic efficiency and personal mobility, safety is a prerequisite of all possible achievements and is becoming the first priority in practice~\citep{du2020online}. In October 2023, Cruise, one of the leading autonomous driving companies, was ordered by California to stop operations of driverless cars in the state after one of Cruise's cars struck a pedestrian in San Francisco~\citep{dara2023california}. The rare incident involved a woman who was first hit by a human driver and then thrown onto the road in front of a Cruise vehicle. The Cruise vehicle then rolled over the pedestrian and finally stopped on top of her, causing serious injuries. Such an accident reflects one of the greatest challenges in autonomous driving: the safety of an autonomous car is largely determined by the ability to detect and react to rare scenarios rather than common normal situations, which have been well considered during development. Although rare in a long-tailed distribution, unusual driving scenarios do happen and can have large impact on driving safety.

To mitigate the impact of abnormal ego behaviors when outside the design domains, a detection system for anomalous driving scenarios is necessary, the output of which can be potentially used as a high-level decision for motion planning. Recently, deep-learning based anomaly detection (AD) algorithms have been widely adopted in robotic applications where rare events are closely related to safety concerns~\citep{chalapathy2019deep}. Many previous works approached the AD problem using supervised learning~\citep{kahn2021land,ji2020multi,ji2022proactive,schreiber2023attentional}. Such methods directly output probabilities of encountering anomalies based on onboard sensory signals by training on binary ground truth labels indicating the presence or absence of anomalies. While labels can be automatically obtained during data collection (e.g., whether or not a mobile robot encounters an anomaly during navigation is indicated by if the human supervisor disengages the autonomy), abundant positive data are required to train a model that can generalize well to unseen scenarios. However, each anomaly in autonomous driving imposes significant risk due to potential accidents and thus is much more expensive than that in other low-risk robotic applications. Furthermore, driving scenarios are complex with diverse road appearances and traffic participants, leading to a large variance of anomalies. It is almost impossible to collect sufficient labeled data for traffic AD. Moreover, there are no guarantees on the completeness of observed anomalies in training data as novel anomaly can always occur in the real world~\citep{liu2018future}.

An alternative solution is unsupervised anomaly detection and/or out-of-distribution detection, which models normal scenarios by training only on negative data without any anomalies. During test time, a high anomaly score, indicating a high probability of anomalies happening, is produced when an observation does not fit the learned distribution. Such unsupervised approaches do not require expensive positive data and can potentially capture any type of anomalous events, even those not previously observed before deployment~\citep{nayak2021comprehensive}. Unsupervised AD has been widely explored in AD tasks for static surveillance cameras~\citep{luo2017revisit,hasan2016learning,morais2019learning,wang2023memory}. However, these methods do not generalize to AD in robotics or autonomous driving as they assume a static camera and thus struggle from rapid camera movement.

\begin{figure}[t]
  \centering
  \begin{subfigure}[b]{\linewidth}
    \captionsetup{justification=centering}
    \includegraphics[width=\linewidth]{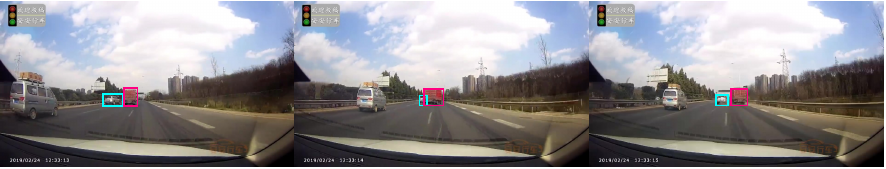}
    \caption{A lateral collision in the distance at high speed.}
    \label{subfig:motivations-small-region}
  \end{subfigure}
  \begin{subfigure}[b]{\linewidth}
    \captionsetup{justification=centering}
    \includegraphics[width=\linewidth]{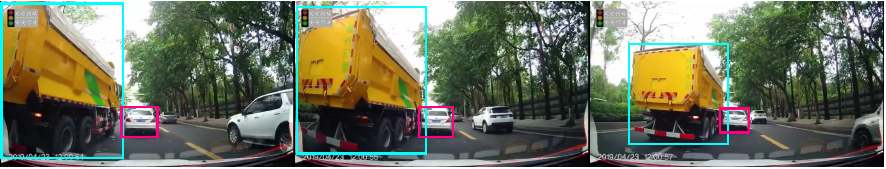}
    \caption{A lateral collision in the proximity at low speed.}
    \label{subfig:motivations-smooth-motion}
  \end{subfigure}
  \caption{\textbf{Challenging examples of anomalies.} Sample frames are ordered in time. Bounding boxes in each frame mark the anomaly participants.}
  \label{fig:motivations}
\end{figure}

In the domain of on-road/traffic AD in egocentric videos, two different classes of approaches are mainly adopted by learning features at different scales. Frame-level methods, which are borrowed from AD for static cameras, train a deep neural network to reconstruct or predict video frames and declare an anomaly whenever the reconstruction or prediction error is large~\citep{liu2018future,hasan2016learning}. However, in addition to the difficulty introduced by moving cameras, anomalous regions can be too small to increase the frame-level error noticeably (Figure~\ref{subfig:motivations-small-region}), leading to miss detection. To overcome this problem, object-centric methods were proposed for on-road AD, which predicts future locations of traffic participants over a short horizon and monitors the accuracy or consistency of the predictions as evidence of anomalies~\citep{yao2019unsupervised}. However, such methods can completely fail in some ego involved anomalies where no other road participants are detected. Furthermore, anomalies featuring mild motions at low speed can also be missed as object's motion appears smooth and can be predicted consistently with small errors (Figure~\ref{subfig:motivations-smooth-motion}). A natural idea is to combine frame-level and object-centric methods to take advantage of the strengths, which has also been explored in recent works with pseudo anomaly score map fusion~\citep{yao2022dota} and cascaded score addition with min-max normalization~\citep{fang2022traffic}. However, how to design and fuse different models effectively in an ensemble remains an open problem for AD in complex driving scenarios.

In this paper, we provide a framework for robot AD module design by performing a holistic analysis on common anomaly patterns. Based on the characteristics of each anomaly type in driving scenarios, we propose three different experts for detecting anomalous scenes, interactions, and behaviors, respectively. The scene expert inherits the advantages of frame-level methods by modeling normal scenes and scene motions and raises an alarm whenever the observed global appearance is out of distribution. Both frame prediction and reconstruction are incorporated to enhance the AD capability. The interaction expert complements existing object-centric methods by focusing on \textit{relative} motions between two road participants and triggers high anomaly scores whenever abnormal interactions (e.g., collisions of cars at both high and low speed) emerge. Inspired by prior works, the behavior expert predicts future locations of objects and monitors the prediction consistency for AD. However, important modifications are made to the existing anomaly score computation to faithfully reflect the probability of an anomalous behavior present in a scene.

To make use of all the modules, we introduce an e\textbf{X}pert \textbf{en}semble, which we call Xen, that fuses anomaly scores from each expert using a Kalman filter. We view each expert score as a \textit{noisy} observation of a ground truth metric quantifying the likelihood of a specific type of anomaly happening and retrieve the final anomaly score as one of the states in the Kalman filter. A new evaluation protocol is also proposed to reflect more realistic performance of AD methods by removing the video-wise min-max normalization of anomaly scores.

Our contributions can be summarized as follows:
\begin{enumerate}
\item
We categorize anomaly patterns for autonomous systems according to the types of involved actors to derive learning objectives of the anomaly detector.
\item
Three experts are proposed to model different normal patterns for autonomous driving (i.e., driving scenes, object interactions, and object behaviors) and to produce high anomaly scores whenever an observation is out of distribution.
\item
We realize expert ensemble through a Kalman filter, one of whose states is the final anomaly score after fusion, to combat against the noise in individual expert scores.
\item
Our proposed detector Xen outperforms existing methods in anomaly detection performance on a large-scale real-world dataset for traffic AD with a newly introduced evaluation protocol for realistic model performance.
\end{enumerate}

%% file: Sections/02-RelatedWork.tex
\section{Related work}
\label{sec:related_work}
Anomaly detection, also known as outlier detection or novelty detection, is an important problem that has been studied within diverse research areas and application domains~\citep{chandola2009anomaly,chalapathy2019deep}. The problem of traffic AD bares similarities with the disciplines of robot AD and AD for surveillance cameras. In this section, we briefly review the related research and introduce common techniques in ensemble deep learning.

Recent research efforts have made noteworthy progress in developing learning-based AD algorithms for robots and mechanical systems. \cite{malhotra2016lstm} introduces an LSTM-based encoder-decoder scheme for multi-sensor AD (EncDec-AD) that learns to reconstruct normal data and uses reconstruction error to detect anomalies. \cite{park2018multimodal} proposes an LSTM-based variational autoencoder (VAE) that fuses sensory signals and reconstructs their expected distribution. The detector then reports an anomaly when a reconstruction-based anomaly score is higher than a state-based threshold. \cite{feng2022unsupervised} attacks multimodal AD with missing sources at any modality. A group of autoencoders (AEs) first restore missing sources to construct complete modalities, and then a skip-connected AE reconstructs the complete signal. Although similar in ideas, these approaches were proposed for low-dimensional signals (e.g., accelerations and pressures) and have not shown effective on high-dimensional data (e.g., images).

AD for robot navigation often involves complex perception signals from cameras and LiDARs in order to understand the environment. \cite{ji2020multi} proposes a supervised VAE (SVAE) model, which utilizes the representational power of VAE for supervised learning tasks, to identify anomalous patterns in 2D LiDAR point clouds during robot navigation. The predictive model proposed in LaND~\citep{kahn2021land} takes as input an image and a sequence of future control actions to predict probabilities of collision for each time step within the prediction horizon. \cite{schreiber2023attentional} further enhances the robot perception capability with the fusion of RGB images and LiDAR point clouds using an attention-based recurrent neural network, achieving improved AD performance on field robots. Different from these supervised-learning-based methods, \cite{wellhausen2020safe} uses normalizing flow models to learn distributions of normal samples of multimodal images, in order to realize safe robot navigation in novel environments. However, driving scenarios have additional complexities than field environments. While road environments are more structured than field environments, additional hazards arise from the presence of and interactions between dynamic road participants, which pose extra challenges on AD algorithms.

Another widely explored research area that is relevant to our work is AD for surveillance cameras, which mainly focuses on detecting the start and end time of anomalous events within a video. Under the category of frame-level methods, \cite{hasan2016learning} proposes a convolutional autoencoder to detect anomalous events by reconstructing stacked images. \cite{chong2017abnormal} and~\cite{luo2017remembering} extend such an idea by learning spatial features and the temporal evolution of the spatial features separately using convolution layers and ConvLSTM layers~\citep{shi2015convolutional}, respectively. Instead of reconstructing frames, \cite{liu2018future} trains a fully convolutional network to predict future frames based on past observations and uses the Peak Signal to Noise Ratio of the predicted frame as the anomaly score. \cite{gong2019memorizing} develops an autoencoder with a memory module, called memory-augmented autoencoder, to limit the generalization capability of the network on reconstructing anomalies. To focus more on small anomalous regions, patch-level methods generate the anomaly score of a frame as the max pooling of patch errors in the image rather than the averaged pixel error used in frame-level methods~\citep{wang2023memory}. In addition, object-level approaches have also been explored, which often focus on modeling normal object motions either through extracted features (e.g., human skeletons)~\citep{morais2019learning} or raw pixel values within bounding boxes~\citep{liu2021hybrid}. Although these methods have achieved promising results on surveillance cameras, the performance is often compromised in egocentric driving scenarios due to moving cameras and complex scenes~\citep{yao2022dota}.

In the domain of traffic AD in first-person videos, pioneering works borrow ideas from surveillance camera applications and detect abnormality by reconstructing motion features at frame level~\citep{yuan2016anomaly}. However, to overcome the issues introduced by rapid motions of cameras and thus backgrounds, object-centric methods are becoming increasingly popular. One of the most representative works by~\cite{yao2019unsupervised} proposes a recurrent encoder-decoder framework to predict future trajectories of an object in the image plane based on the object's past trajectories, spatiotemporal features, and ego motions. The accuracy and consistency of the predictions are then used to generate anomaly scores. One critical problem of such a method is the inevitable miss detection in the absence of traffic participants. As a result, ensemble methods emerge recently to combine the strengths of frame-level and object-centric methods. For example, ~\cite{yao2022dota} fuses the object location prediction model with the frame prediction model to achieve all-scenario detection capability and ~\cite{fang2022traffic} monitors the temporal consistency of frames, object locations, and spatial relation structures of scenes for AD. In this work, we derive each module and the corresponding learning objective in the ensemble based on a comprehensive analysis on anomaly patterns in egocentric driving videos. In particular, an interaction module is introduced to monitor anomalous interactions between road participants. The scores from each module are then fed as observations of a Kalman filter, from which the final anomaly score is obtained.

In this paper, we introduce an ensemble of detectors to capture different classes of anomalies.
We take inspiration from recent advances in ensemble deep learning, which aims to improve the generalization performance of a learning system by combing several individual deep learning models~\citep{ganaie2022ensemble} and has been applied to different application domains, such as speech recognition~\citep{li2017semi},  image classification~\citep{wang2020particle}, forecasting~\citep{singla2022ensemble}, and fault diagnosis~\citep{wen2022new}. Out of different classes of ensemble deep learning approaches, the most similar work to Xen is the heterogeneous ensemble (HEE), in which the components are trained on the same dataset but use different algorithms/architectures~\citep{li2018heterogeneous,tabik2020mnist}. However, each component in an HEE is usually trained with an identical learning objective, while each expert in Xen is assigned with different learning tasks. In terms of result fusion strategies, unweighted model averaging is one of the most popular approaches in the literature~\citep{ganaie2022ensemble}, which simply averages the outcomes of the base learners to get the final prediction of the ensemble model. By contrast, we exploit a Kalman filter to further combat the noise in scores from different components in a time-series task, and the unweighted model averaging can be viewed as a special case of such a method at a point along the time axis.

%% file: Sections/03-Overview.tex
\section{Problem overview}
\label{sec:overview}

Our goal is to develop an AD module that enables an autonomous car to detect anomalous events online in diverse driving scenarios.

\begin{figure*}[t]
  \centering
  \begin{subfigure}[b]{0.35\linewidth}
    \captionsetup{justification=centering}
    \includegraphics[width=\linewidth]{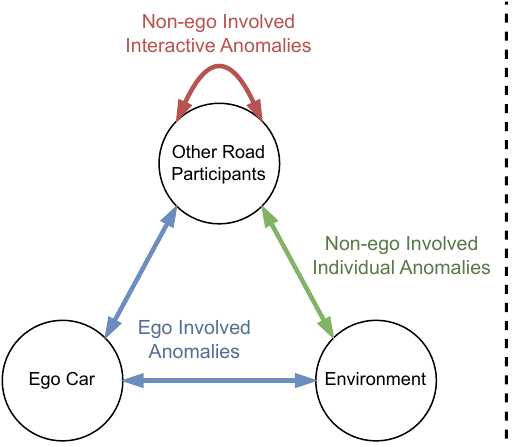}
    \caption{Common anomaly patterns in driving.}
    \label{fig:anomaly-patterns}
  \end{subfigure}
  \begin{subfigure}[b]{0.64\linewidth}
    \captionsetup{justification=centering}
    \includegraphics[width=\linewidth]{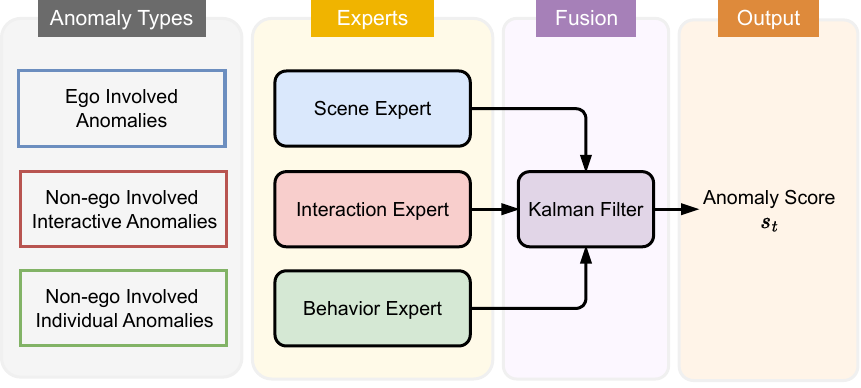}
    \caption{System structure overview.}
    \label{fig:Xen-overview}
  \end{subfigure}
  \caption{\textit{Left:} Nodes are possible actors in an on-road anomaly. Each edge represents a type of anomaly happening between the two connected nodes. The edges are grouped by colors into three categories to guide our anomaly detector design. \textit{Right:} Three experts are proposed for detecting different types of anomalies based on the anomaly pattern analysis. Individual expert scores are fused by a Kalman filter to generate a comprehensive final score.}
  \label{fig:problem-overview}
\end{figure*}

To guide our anomaly detector design, an analysis of common anomaly patterns in driving scenarios is presented in Figure~\ref{fig:anomaly-patterns}. In most of the anomalous events on road, there exist three main players, namely the ego car, other road participants (e.g., cars, motorcycles, pedestrians), and the environment (e.g., guardrails, traffic signs, slippery roads), which are represented as the graph nodes. Any two of the nodes can produce a type of anomaly: the edge between the ego car and other road participants includes abnormal cases such as a collision between the ego car and a pedestrian; the edge between the ego car and the environment encompasses accidents such as the ego car being out of control due to slippery roads after raining; the edge connecting other road participants and the environment stands for \textit{non-ego involved individual anomalies} where a \textit{single} agent behaves abnormally, such as a car colliding with a guardrail. In particular, there is a self-loop around the node of other road participants, which represents \textit{non-ego involved interactive anomalies} where anomalous \textit{interactions} happen between other agents, such as two vehicles colliding with each other. We further summarize the two edges with one of the ends being the ego car as \textit{ego involved anomalies}.

Note that the presented graph is general enough to encompass rare moments with more than one anomaly present by activating multiple edges simultaneously. For example, in an event where two cars collide with each other and another car swerves due to loss of control, both \textit{non-ego involved interactive anomalies} and \textit{non-ego involved individual anomalies} edge will be activated. Moreover, two \textit{non-ego involved individual anomalies} edges will be activated if two non-ego cars lose control \textit{independently} in the same time. The AD problem can now be converted into activating relevant graph edges properly and promptly in abnormal events. Note that our AD problem formulation is not limited to automated cars and can be generalized to other robotic applications with proper modifications.

We approach the AD task in autonomous driving using unsupervised learning to avoid the difficulty and cost of collecting large-scale labeled anomaly data for training. Different normal patterns in non-anomalous driving videos have been modeled in prior works to detect out-of-distribution samples during test time for AD. However, learning unimodal normal patterns can only capture a subset of the edges in Figure~\ref{fig:anomaly-patterns}. For example, frame prediction framework is good at detecting \textit{ego involved anomalies} as large unexpected changes in pixel values are often observed in such events but can miss some of the non-ego involved anomalies due to the small occupancy of anomalous objects in the scene~\citep{liu2018future,hasan2016learning}. Object trajectory prediction approach can capture \textit{non-ego involved individual anomalies} well but can miss \textit{non-ego involved interactive anomalies} when the behaviors of both anomaly participants appear smooth individually~\citep{yao2019unsupervised,yao2022dota}. To cover the entire graph in Figure~\ref{fig:anomaly-patterns}, an ensemble of different experts for different edges are necessary.

Formally, we assume that the observations available at time $t$, $\mathbf{o}_t$, consist of the current and past RGB images $\mathbf{o}_t \coloneqq \left( I_1, I_2, \dots, I_t \right)$ with $I_i \in \mathbb{R}^{H \times W \times 3}$ from a forward facing monocular camera. At each time step, the task for the AD module, denoted as the function $g$, is to regress an \textit{anomaly score} $s_t \in \mathbb{R}$ from the current available observations $\mathbf{o}_t$: $s_t = g(\mathbf{o}_t)$. A higher anomaly score indicates a higher probability of an anomaly happening. To ensure the generalization performance of our anomaly detector, we make no assumptions on the consistency of camera intrinsics nor camera poses across videos.

The system structure of Xen is shown in Figure~\ref{fig:Xen-overview}. Compared to the existing AD methods for robotic systems, Xen is able to build upon the holistic analysis of anomaly patterns to design different experts for different types of anomalous events. Specifically, Xen consists of:
\begin{enumerate*}[label=(\arabic*)]
\item
a \textit{scene expert} modeling normal scenes and scene motions at frame level for detecting \textit{ego involved anomalies},
\item
an \textit{interaction expert} modeling normal relative motions between two objects for detecting \textit{non-ego involved interactive anomalies}, and
\item
a \textit{behavior expert} modeling normal object trajectories for detecting \textit{non-ego involved individual anomalies}.
\end{enumerate*}
An ensemble function $e$, which is realized by a Kalman filter, is then applied to fuse the scores from all the experts for enhanced performance:
\begin{equation*}
s_t = e(g_\text{s}(\mathbf{o}_t), \, g_\text{i}(\mathbf{o}_t), \, g_\text{b}(\mathbf{o}_t)),
\end{equation*}
where $g_\text{s}$, $g_\text{i}$, and $g_\text{b}$ are AD experts for modeling normal scenes, interactions, and behaviors, respectively. Such a multi-expert design can also enable the \textit{classification} of an anomaly by comparing the scores from the experts, even though no labels were used during training. Lastly, we modify the common evaluation protocol for AD tasks by removing video-wise anomaly score normalization so that the model performance in the experiments is more realistic as a real-world system (Section~\ref{sec:experiments}).

In the following sections, we describe the design of the three experts $g_\text{s}$, $g_\text{i}$, and $g_\text{b}$ for different anomaly types and the ensemble function $e$ for the improved overall performance than each individual component.

%% file: Sections/04-Experts.tex
\section{Anomaly detection experts in autonomous driving}

In this section, we present the model architecture, training objective, and anomaly score generation of each expert. Specifically, \textit{scene}, \textit{interaction}, and \textit{behavior} experts are introduced for \textit{ego involved}, \textit{non-ego involved interactive}, and \textit{non-ego involved individual} anomalies, respectively, as identified in Section~\ref{sec:overview}. Targeting online AD, our overall design philosophy is to keep each module parallelizable to one another and easy to implement. All the experts are trained independently with a large-scale dataset of normal driving videos without any anomalies.

\subsection{Scene expert}

Some of the on-road anomalies are often accompanied with large unexpected scene changes in egocentric videos, especially for \textit{ego involved anomalies}. In events where the ego dynamics deviate significantly from the nominal one (e.g., when the ego car swerves, rotates suddenly due to a side collision, is rear-ended), future frames become unpredictable from the perspective of the ego camera, leading to a large prediction error in video frames. However, false negatives can still happen if we only rely on detecting large inconsistency across frames. For example, when the ego car slips forward slowly and hits another car in the back, future frames can still be predicted accurately due to the slow motion of the car, making the system refuse to raise an alert. A useful hint for anomaly detection in such a case is the rarity of the scene itself: the preceding car would never be this close to the ego camera (spatial features) and keep getting closer (temporal features) in normal scenarios.

In order to detect \textit{ego involved anomalies} in both cases of significant dynamics change and slow motions, we assemble the frame-level \textit{scene expert} with two submodules: future frame prediction (FFP) and spatial-temporal reconstruction (STR).

\subsubsection{Future frame prediction.}
FFP aims to detect anomalies by monitoring large differences between the predicted future frame and the actual future frame. The preceding $T_\text{ffp}$ frames are used for prediction.

\begin{figure}[t]
  \centering
  \includegraphics[width=\linewidth]{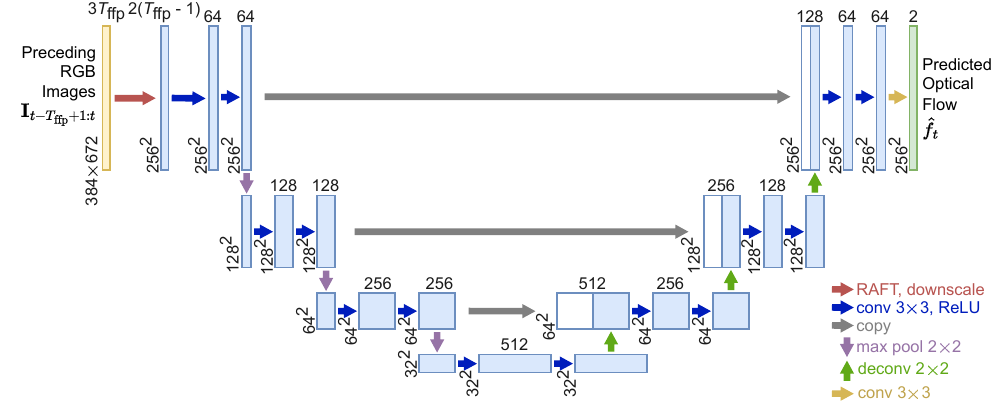}
  \caption{\textbf{Model architecture of OFP.} Each box corresponds to a multi-channel feature map. The number of channels is denoted on top of the box. The spatial resolution is provided at the lower left edge of the box. Arrows with different colors denote different operations.}
  \label{fig:ofp}
\end{figure}

It is known that optical flow information plays an important role in predicting future images: given the image $I_t$ and the optical flow $f_t$ between $I_t$ and $I_{t+1}$, the future image $I_{t+1}$ can be predicted accurately by shifting pixels~\citep{fang2022traffic}. However, $f_t$ is unknown at time $t$. A common solution is to utilize the optical flow information retrospectively during training: the predicted image $\hat{I}_{t+1}$ is first generated, then the difference between the induced optical flow and the ground truth optical flow at time $t$ is incorporated into the loss function for optimization~\citep{liu2018future,ye2019anopcn}. Alternatively, we explicitly predict the scene motion $\hat{f}_t$, which will also be utilized in STR, before frame prediction. The network structure of such an optical flow prediction (OFP) module is shown in Figure~\ref{fig:ofp}. The preceding $T_\text{ffp}$ RGB images $\mathbf{I}_{t - T_\text{ffp} + 1:t} \coloneqq (I_{t - T_\text{ffp} + 1}, I_{t - T_\text{ffp} + 2}, \cdots, I_t)$ are first processed by pretrained RAFT~\citep{teed2020raft}, an optical flow estimation network, to generate scene motions $\mathbf{f}_{t - T_\text{ffp} + 1:t - 1} \coloneqq (f_{t - T_\text{ffp} + 1}, f_{t - T_\text{ffp} + 1}, \cdots, f_{t - 1})$ in the past $T_\text{ffp} - 1$ frames. We perform RAFT inference at a relatively high resolution of $384 \times 672$ to ensure the quality of the estimated optical flow, which is then downsampled to $256 \times 256$ for fast training\endnote{Each entry value in the predicted optical flow should also be scaled proportionally to the ratio change of downsampling to maintain the validity of the flow at the new resolution.}. The resulting optical flows $\mathbf{f}_{t - T_\text{ffp} + 1:t - 1}$ in pixel units are then concatenated temporally and fed as the input to subsequent layers of OFP, which are based on u-net~\citep{ronneberger2015u}, to predict the current scene motion $\hat{f}_t$. The resolution of each feature map at the same level stays unchanged by using a padding of 1 in each convolution layer to avoid the cropping operation in the u-net paper. To reach a balance between the computational complexity and performance, we set $T_\text{ffp}$ as 4, similar to prior works~\citep{liu2018future,fang2022traffic}.

\begin{figure}[t]
  \centering
  \includegraphics[width=\linewidth]{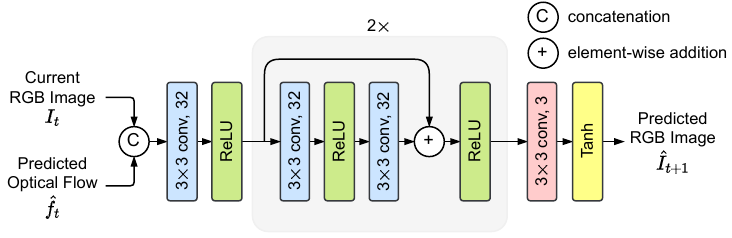}
  \caption{\textbf{Model architecture of FFP.} Each circle and box represents an operation. The last number in the box of each convolution layer denotes the number of filters.}
  \label{fig:ffp}
\end{figure}

Upon obtaining $\hat{f}_t$ from OFP, the future frame can be predicted by combining $I_t$ and $\hat{f}_t$. The details of FFP are illustrated in Figure~\ref{fig:ffp}. The channel-wise concatenation of the current downscaled RGB image and the predicted optical flow are processed sequentially by a convolution layer, two residual convolution blocks, and a final convolution layer to generate a future RGB image $\hat{I}_{t+1}$. The spatial resolution of all the feature maps, including the input and output, within FFP is kept as $256 \times 256$. Tanh is used as the final activation function as the pixel values in images are normalized to $[-1, 1]$.

We train both networks jointly in one stage. To make the prediction close to the ground truth, $L_2$ and gradient loss are used, following prior works~\citep{liu2018future}. The $L_2$ loss aims to guarantee the similarity of all pixels in RGB space by penalizing the distance between a predicted frame $\hat{I}$ and the corresponding actual frame $I$:
\begin{equation}
L_2(\hat{I}, I) = \| \hat{I} - I \|_2^2.
\end{equation}
The gradient loss encourages a sharp synthetic image with the form of:
\begin{equation}
\begin{aligned}
L_\text{grad}(\hat{I}, I) = \sum_{i,j,c} &\left|| \hat{I}_{i,j,c} - \hat{I}_{i-1,j,c} | - | I_{i,j,c} - I_{i-1,j,c} |\right| \\
+
&\left|| \hat{I}_{i,j,c} - \hat{I}_{i,j-1,c} | - | I_{i,j,c} - I_{i,j-1,c} |\right|,
\end{aligned}
\end{equation}
where $i,j,c$ denote the spatial index of an element within an RGB image. In addition, we introduce an optical flow loss based on smooth $L_1$ loss~\citep{girshick2015fast} to supervise the learning of the intermediate optical flow:
\begin{equation}
L_\text{of} (\hat{f}, f) = \text{smooth}_{L_1} (\hat{f} - f),
\end{equation}
where $\hat{f}$ is the output of OFP, and $f$ is given by applying RAFT on the two corresponding ground truth images.

The overall training objective of FFP is a an addition of the intensity, gradient, and optical flow loss:
\begin{equation}
\begin{aligned}
L_\text{ffp} = L_2 (\hat{I}_{t+1}, I_{t+1}) + L_\text{grad} (\hat{I}_{t+1}, I_{t+1}) + L_\text{of} (\hat{f}_t, f_t).
\end{aligned}
\end{equation}
Although the relative weights between the three losses can be further tuned, we choose an identical value for all the terms for simplicity.

We denote the anomaly detector using a trained FFP as a function $g_\text{ffp}: \mathbf{I}_{t - T_\text{ffp}:t} \mapsto s_t^\text{ffp}$, which takes as input $T_\text{ffp} + 1$ images and outputs an anomaly score $s_t^\text{ffp}$ for time $t$. Specifically, the first $T_\text{ffp}$ images are used to produce the predicted frame $\hat{I}_t$, and the anomaly score is calculated as the negative Peak Signal to Noise Ratio (PSNR):
\begin{equation}
s_t^\text{ffp} = -\text{PSNR}(\hat{I}_t, I_t) = 10\log_{10} \left(\frac{1}{M} \| \hat{I}_t - I_t \|_2^2\right),
\end{equation}
where $M$ is the number of pixels in an image, and $\hat{I}_t$ and $I_t$ are normalized to $[0, 1]$ before anomaly score computation. A higher anomaly score $s_t^\text{ffp}$ (i.e., lower PSNR) indicates a higher probability of an anomaly present at time $t$.

\subsubsection{Spatial temporal reconstruction.} Reconstructing sensor signals is another pervasive approach in unsupervised AD as unseen scenarios can be directly monitored and detected. Frame reconstruction can complement frame prediction in cases where the anomaly is not accompanied with large inconsistency across frames, such as a slight rear-end collision due to slipping or scratches with guardrails due to gradual deviation from the lane center.

One of the key design choices in scene reconstruction is the input space. A natural idea is to represent a scene in RGB or grayscale values~\citep{luo2017remembering,hasan2016learning}. However, in complex driving scenarios, the intra-class variance of samples using such a representation is high, thus leading to frequent false alarms. For example, if there is a colorful car for some special events driving in front of the ego car, the scene is likely to be categorized as an anomaly as such a car painting has never been observed in normal training data, although the scene should belong to normal cases as all the driving-related events appear normal. As a result, we express driving scenes in depth and motion space, delivering spatial and temporal features of a scene, respectively. Such an input space can selectively ignore novel appearances of normal objects and focus more on cues that indeed lead to anomalous events.

\begin{figure}[t]
  \centering
  \includegraphics[width=\linewidth]{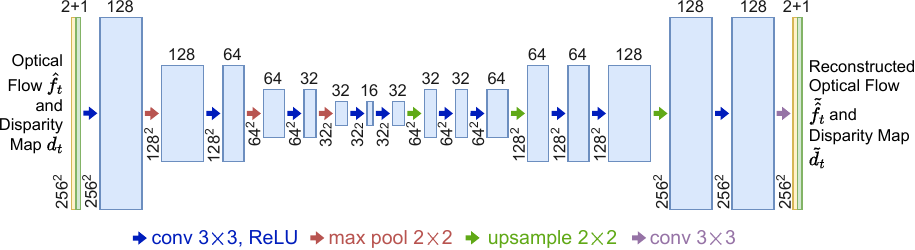}
  \caption{\textbf{Model architecture of STR.} The optical flow and disparity map are concatenated and reconstructed by a fully convolutional autoencoder. The annotations can be interpreted in the same way as those in Figure~\ref{fig:ofp}.}
  \label{fig:str}
\end{figure}

STR is designed to detect anomalies by observing the reconstruction error of the spatiotemporal features of a scene. Our network structure based on Conv-AE~\citep{hasan2016learning} is shown in Figure~\ref{fig:str}. Although the scene motion $f_t$ is unknown at time $t$, an estimation $\hat{f}_t$ can be predicted from $\mathbf{I}_{t - T_\text{ffp} + 1:t}$ by OFP. We further generate a disparity map $d_t$ (i.e., inverse depth map up to scale) of the scene $I_t$ using pretrained MiDaS~\citep{ranftl2020towards}. Due to the scale ambiguity in monocular depth estimation, a min-max normalization on disparity values within an image is often performed~\citep{ranftl2020towards,godard2019digging,godard2017unsupervised}. However, such a \textit{relative} depth tends to eliminate informative distinctions between frames. For example, consider a scene with only a car close to the ego car and another scene with only a distant car. The disparity values of these two cars will be identical after normalization even though the metric depths are different. Therefore, we remove the min-max normalization to preserve global distinctions and assume that the estimated disparity and the actual metric depth are correlated. Disparity values are then clipped and divided by an upper bound globally on the entire dataset for a normalization to $[0, 1]$.

The concatenation of the predicted optical flow and the disparity map is then processed by a convolutional autoencoder. The bottleneck in the middle of the network, with the fewest channels and lowest resolution among all the feature maps, ensures that only representative common features of normal scenes are extracted and that the reconstruction will be of low quality if such normal patterns are absent. Although memory-augmented AE has been shown effective in surveillance~\citep{gong2019memorizing,park2020learning}, we choose a simple architecture since the diversity in egocentric driving videos can require a significant amount of memory. Notably, we replace each deconvolution layer in Conv-AE with a nearest-neighbor interpolation followed by two convolution layers to counteract the checkerboard artifacts in reconstructed signals~\citep{odena2016deconvolution}. Similar to OFP and FFP, a padding of 1 is used to maintain the resolution for each convolution layer.

\begin{figure*}[t]
  \centering
  \includegraphics[width=0.85\linewidth]{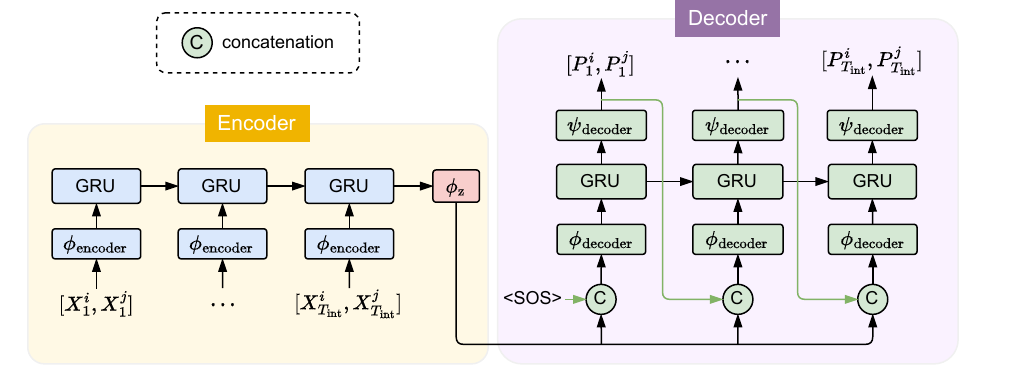}
  \caption{\textbf{Model architecture of the interaction expert.} The network follows an encoder-decoder structure, where a pair of trajectories is compressed into a low-dimensional space and then decoded to generate the parameters for trajectory pair reconstruction. The start-of-sequence state is denoted by $\langle \text{SOS} \rangle$. The hidden state of both GRUs are initialized to zeros.}
  \label{fig:im}
\end{figure*}

The OFP parameters are frozen during the training of STR. An $L_1$ loss and a total variation loss are used to supervise the reconstruction of the original signal and combat further against the checkerboard artifacts for smoothness, respectively:
\begin{gather}
L_1 (\tilde{s}, s) = \| \tilde{s} - s \|_1, \\
L_\text{tv} (\tilde{s}) = \sum_{i,j,c} \left(\tilde{s}_{i,j,c} - \tilde{s}_{i-1,j,c}\right)^2 + \left(\tilde{s}_{i,j,c} - \tilde{s}_{i,j-1,c}\right)^2,
\end{gather}
where $s$ and $\tilde{s}$ are the original and reconstructed signal, respectively. The overall training objective of STR can be expressed as:
\begin{equation}
\label{eq:str-objective}
\begin{aligned}
L_\text{str}
=
&\left(
\lambda_\text{d} L_1 (\tilde{d}_t, d_t)
+
L_1 (\tilde{\hat{f}}_t, \hat{f}_t)
\right) \\
&+
\lambda_\text{tv}
\left(
\lambda_\text{d} L_\text{tv}(\tilde{d}_t)
+
L_\text{tv} (\tilde{\hat{f}}_t)
\right),
\end{aligned}
\end{equation}
where $\lambda_\text{d}$ is a scalar controlling the relative weight between the loss on the disparity and optical flow map and $\lambda_\text{tv}$ is a scalar balancing the $L_1$ and total variation loss. During evaluation, the anomaly detector using STR can be viewed as a function $g_\text{str}:\mathbf{I}_{t-T_\text{ffp}+1:t} \mapsto s_t^\text{str}$, in which the anomaly score $s_t^\text{str}$ is set to be the same as $L_\text{str}$ at time $t$.

With FFP and STR, our scene expert offers two anomaly scores based on frame prediction and reconstruction:
\begin{equation}
g_\text{s} (\mathbf{o}_t)
=
\{ g_\text{ffp} (\mathbf{I}_{t - T_\text{ffp}:t}), g_\text{str} (\mathbf{I}_{t - T_\text{ffp}+1:t}) \}
=
\{s_t^\text{ffp}, s_t^\text{str}\}.
\end{equation}
As can be seen, FFP starts producing anomaly score $s_t^\text{ffp}$ at time $T_\text{ffp} + 1$ while STR starts producing anomaly score $s_t^\text{str}$ at time $T_\text{ffp}$ in a video.

\subsection{Interaction expert}
\label{subsec:im}
Although the scene expert is good at detecting \textit{ego-involved anomalies} by learning frame-level normal patterns, non-ego involved anomalies can be easily missed due to the relatively small scales of anomaly participants in frames. Some object-level anomalies are often accompanied with abnormal \textit{interactions}, which we define as the relative motion between two objects. For example, a collision between two cars is often preceded by a shrinking relative distance without sufficient deceleration, which is not a common interaction pattern in normal driving scenarios. An anomaly detector modeling interactions can also potentially raise alarms earlier than that modeling individual behaviors as anomalous interactions (e.g., approaching at high relative velocity) appear earlier than anomalous individual behaviors (e.g., sudden stop of a car after a collision).

Our interaction expert models normal interactions by reconstructing motions of two objects, and an anomaly is detected whenever a high reconstruction error is observed. We denote a detected object's bounding box as $X_t=[c_t^x, c_t^y, w_t, h_t]$, where $[c_t^x, c_t^y]$ is the location of the center of the box and $w_t$ and $h_t$ are the width and height of the box, respectively. Given a pair of bounding box trajectories in the image plane $\{X_{t - T_\text{int} + 1:t}^i, X_{t - T_\text{int} + 1:t}^j\}$, where $T_\text{int}$ is the history horizon of the interaction expert and $i,j$ denote the IDs of the two objects, our time series model reconstructs the two trajectories simultaneously. We use pretrained Mask R-CNN~\citep{he2017mask}, following prior works~\citep{yao2022dota,yao2020and,fang2022traffic}, and StrongSORT~\citep{du2023strongsort} to acquire and associate object bounding boxes across frames at the original resolution, respectively. Each bounding box is normalized with respect to the dimensions of the images to facilitate efficient training.

Our gated recurrent unit based autoencoder (GRU-AE) for trajectory pair reconstruction, shown in Figure~\ref{fig:im}, is inspired by the work on driver trait modeling~\citep{liu2022learning}. For brevity, we denote each pair of trajectories fed as input to GRU-AE at time $t$ as a one-indexed sequence of length $T_\text{int}$, as shown in the input and output in Figure~\ref{fig:im}. Each pair of bounding boxes at time $k$ within a trajectory pair is flattened into a 1D vector to represent the joint localization of the two objects. The encoder GRU then applies a non-linear embedding function $\phi_\text{encoder}$ to each flattened pair of bounding boxes and feeds the embedded features to the GRU cell:
\begin{equation}
\label{eq:im-encoder}
h_k^\text{e} = \text{GRU} \left( h_{k-1}^\text{e}, \phi_\text{encoder} ([X_k^i, X_k^j]) \right),
\end{equation}
where $h_k^\text{e}$ is the hidden state of the encoder GRU at time $k \in (1, \dots, T_\text{int})$. After the entire trajectory pair is processed, we take the last hidden state $h_{T_\text{int}}^e$ as the feature of the interaction and create a bottleneck through another fully connected layer $\phi_\text{z}$ to distill representative normal patterns:
\begin{equation}
z = \phi_\text{z} (h_{T_\text{int}}^\text{e}).
\end{equation}
We choose an AE over a variational autoencoder (VAE) since we observe that any regularization of the latent space significantly degrades the reconstruction performance of the network in our experiments, possibly due to the highly diverse interaction patterns in normal driving videos.

GRU-AE is tasked with reconstructing the original bounding box trajectories. However, directly outputting the absolute coordinates of bounding boxes from the network is inefficient to achieve our goal: we care more about object \textit{motions} than object \textit{positions}. In other words, it is the motion that the two cars are approaching each other, not where the two cars collide in the frame, that leads to an anomaly. Therefore, we use the first pair of bounding boxes in the sequence as two anchor boxes, similar to the idea in object detection~\citep{redmon2018yolov3}, and the network only outputs the parameters $[P_k^i, P_k^j]$ used for transforming the anchor boxes into target boxes. More specifically, for an object (ID omitted for brevity), the parameter $P_k$ at time $k$ consists of four elements $p_k^x$, $p_k^y$, $p_k^w$, $p_k^h$ for the transformation:
\begin{equation}
\label{eq:new-box-from-anchor-box}
\begin{aligned}
\tilde{c}_k^x &= c_1^x + p_k^x, \\
\tilde{c}_k^y &= c_1^y + p_k^y, \\
\tilde{w}_k &= w_1 e^{p_k^w}, \\
\tilde{h}_k &= h_1 e^{p_k^h},
\end{aligned}
\end{equation}
where $\tilde{X}_k=[\tilde{c}_k^x, \tilde{c}_k^y, \tilde{w}_k, \tilde{h}_k]$ is the reconstructed bounding box at time $k$ and $X_1=[c_1^x, c_1^y, w_1, h_1]$ is the anchor box. Such a characterization allows the network to focus more on the modeling of motions than absolute positions of bounding boxes, which provide little information on the normality of the scene\endnote{The parameters at the first time step output by a properly trained GRU-AE should always be around the origin as the target at the first time step is just the anchor box.}.

We assume that the interaction pattern stays constant over a small window along a trajectory. As a result, in the decoding stage, the latent state $z$ is concatenated with the decoder output at the previous time step $[P_{k-1}^i, P_{k-1}^j]$ to form a joint state at time $k$, instead of being used to initialize the decoder GRU. Such a joint state is then embedded by a function $\phi_\text{decoder}$, fed into the decoder GRU cell, and processed by another non-linear function $\psi_\text{decoder}$ to generate the parameters for reconstruction at time $k$:
\begin{equation}
\label{eq:im-decoder}
\begin{aligned}
&h_k^\text{d} = \text{GRU} \left( h_{k-1}^\text{d}, \phi_\text{decoder} ([P_{k-1}^i, P_{k-1}^j, z]) \right), \\
&[P_k^i, P_k^j] = \psi_\text{decoder} (h_k^\text{d}),
\end{aligned}
\end{equation}
where $h_k^\text{d}$ is the hidden state of the decoder GRU at time $k \in (1, \dots, T_\text{int})$. At the first time step, we use a special start-of-sequence (SOS) state, similar to the start-of-sequence symbol in natural language processing~\citep{grefenstette2015learning}, to generate $[P_1^i, P_1^j]$. The steps in~(\ref{eq:im-decoder}) are repeated until the parameters for the entire trajectory are generated.

The interaction expert is designed for non-ego involved anomalies. Therefore, the network should pay more attention to small-scale traffic participants as the anomalies associated with large objects are likely to be detected by the scene expert. To this end, instead of performing explicit data mining to filter out trajectory pairs with large bounding boxes, we incorporate our emphasis on small objects into the objective function. Furthermore, to combat against the input noise introduced by the object detection and tracking algorithms, we weight the loss based on the standard deviation of the bounding box coordinates over time (i.e., we trust stable detection more than the unstable one). As a result, the objective for training our GRU-AE is a scaled root mean squared error (RMSE) between the original and reconstructed bounding boxes within a trajectory \textit{pair}:
\begin{equation}
\label{eq:im-loss}
L_\text{int}^{ij} = \sum_{id \in \{i,j\}} \sum_{k=1}^{T_\text{int}} \left( \lambda_\text{h}^{-1} \lambda_\text{std}^{-1} \| \tilde{X}_k^{id} - X_k^{id} \|_2^2 \right)^\frac{1}{2},
\end{equation}
where $\lambda_\text{h}$ is the average height of the original bounding boxes of the two objects, serving as a proxy of the distance from the ego car to the objects, and $\lambda_\text{std}$ is the average standard deviation (STD) of the original bounding box coordinates of the two objects within the time window:
\begin{equation}
\begin{aligned}
\lambda_\text{h} &= \frac{1}{2T_\text{int}} \sum_{id \in \{i,j\}} \sum_{k=1}^{T_\text{int}} h_k^{id}, \\
\lambda_\text{std} &= \frac{1}{8} \sum_{id \in \{i,j\}} \sum_{y \in \{c^x,c^y,w,h\}} \text{STD} (y_{1:T_\text{int}}^{id}).
\end{aligned}
\end{equation}
We further clip $\lambda_\text{std}$ with a lower-bound $\tau_\text{std} > 0$ for numerical stability. With the introduced scaled loss, GRU-AE is able to focus more on the interactions of distant objects and thus better complement the scene expert.

Notably, the number of \textit{pairs} of objects grows quadratically with the number of objects. To set an upper limit on the computational complexity of the interaction expert, we establish and maintain a priority queue within each time window with capacity $N_\text{max}$ to store pairs of trajectories that are most likely to have interactions between each other based on the relative distance of the bounding boxes before the forward pass of GRU-AE. More specifically, each pair of trajectories is associated with a \textit{distance score}, defined as the proximity of the two bounding boxes with the consideration of the box size:
\begin{equation}
\begin{aligned}
ds^{ij} = \min_{1 \le k \le T_\text{int}} &|c_k^{x,i} - c_k^{x,j}| - (w_k^i + w_k^j)/2 \\
+ &|c_k^{y,i} - c_k^{y,j}| - (h_k^i + h_k^j)/2.
\end{aligned}
\end{equation}
A lower \textit{distance score} means that the two objects are closer to each other and thus are more likely to produce an interactive anomaly. Within a time window, the priority queue maintains at most $N_\text{max}$ pairs of trajectories with the lowest \textit{distance score}, which are then fed into GRU-AE for anomaly detection.

\begin{figure*}[t]
  \centering
  \includegraphics[width=0.85\linewidth]{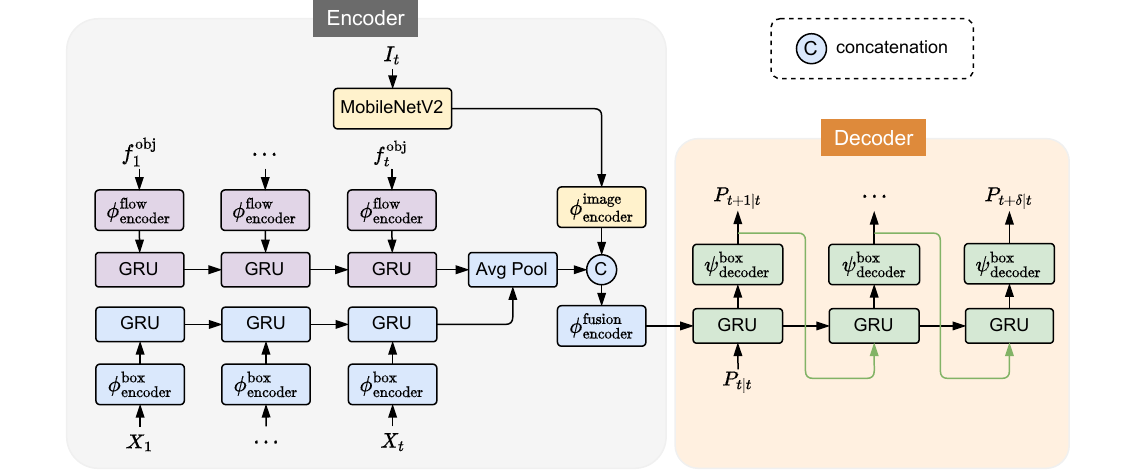}
  \caption{\textbf{Model architecture of the behavior expert.} The object's bounding box, object's optical flow features, and image are fed to different encoders, then fused and decoded to generate the parameters for transforming the anchor box into the predicted bounding boxes within the prediction horizon. The processing pipelines for the bounding box, optical flow features, and image are in blue, purple, and yellow, respectively. The hidden state of the GRUs in the encoder is initialized to zeros.}
  \label{fig:bm}
\end{figure*}

We follow common practice in AE-based anomaly detection by calculating the anomaly score in the same way as the training objective~(\ref{eq:im-loss}) for each pair of trajectories. During evaluation on a video, we denote $PQ_t$ as the set of pairs of trajectories stored in the priority queue that are present from time $t - T_\text{int} + 1$ to $t$ inclusively. The anomaly score $s_t^\text{int}$ at time $t$, generated by the interaction expert, is the average $L_\text{int}$ value over all pairs of trajectories in $PQ_t$:
\begin{equation}
s_t^\text{int} = \frac{1}{|PQ_t|} \sum_{ij \in PQ_t} L_\text{int}^{ij},
\end{equation}
where $|PQ_t|$ is the cardinality of $PQ_t$ that is at most $N_\text{max}$. If no valid pairs of trajectories are detected from time $t - T_\text{int} + 1$ to $t$, $s_t^\text{int}$ is set to be $0$. To compensate for possible discontinuities in the anomaly score $s_t^\text{int}$ over time due to missing objects or incorrect ID associations in the detection and tracking algorithms, we further apply a digital low-pass filter on $s_t^\text{int}$ for smoothness. Such a filter is causal and thus can be utilized for online AD.

\subsection{Behavior expert}
\label{subsec:bm}
In cases where a non-ego involved anomaly only contains a single agent, the interaction expert can fail easily. For example, if there is only one vehicle visible in the field of view and starts swerving due to the snow, the interaction expert will not be activated as no interactions between road participants exist. A useful hint for anomaly in such a case is the observed unusual behavior: a car would never zigzag aggressively in normal driving scenarios.

Our behavior expert models normal behaviors by predicting future locations of a road participant, and an anomaly is detected whenever an object's observed trajectory deviates significantly from the predicted trajectory. In addition to the past bounding boxes of an object, prior works also include ego motions as one of the inputs to compensate for the object location change due to such motions~\citep{yao2019unsupervised,yao2022dota}. However, Monocular SLAM algorithms~\citep{mur2015orb,mur2017orb,campos2021orb}, used to estimate ego motions, require known camera intrinsics, which is often unavailable in a large-scale dataset with videos from multiple sources. Moreover, the tracking is often lost in featureless scenes, such as a highway in a rural area in a plain. Therefore, we remove the ego motions from our inputs to facilitate a more robust system.

Another useful hint for the prediction of future object localization is the global information contained in an image. For example, a normal trajectory of a vehicle is often constrained by obstacle-free drivable areas, within which the predicted future locations should reside. As a result, we made our behavior expert aware of the environment by incorporating the latest full RGB image into the inputs. Furthermore, we include the past optical flow features of each object as a final part of our inputs to enhance the perception on object motions, similar to prior works~\citep{fang2022traffic,yao2022dota}.

For each object, given the past bounding box trajectory $(X_1, X_2, \cdots, X_{t})$, the past optical flow features $(f_1^\text{obj}, f_2^\text{obj}, \cdots, f_t^\text{obj})$, and the original image $I_t$, our future object localization model predicts a trajectory $(\hat{X}_{t+1|t}, \hat{X}_{t+2|t}, \cdots, \hat{X}_{t+\delta|t})$ based on the information up until time $t$, where $\delta > 0$ is the prediction horizon. Each bounding box is normalized with respect to the dimensions of the images as in the interaction expert, and the object's optical flow features are extracted by a $5 \times 5$ RoIAlign~\citep{he2017mask} operation from the frame-level optical flow fields. Our time series model, shown in Figure~\ref{fig:bm}, follows an encoder-decoder structure, similar to GRU-AE proposed in Section~\ref{subsec:im} but without a bottleneck. One major difference between the behavior and interaction expert is that the behavior expert only needs to process the latest data (i.e., $X_t$, $f_t^\text{obj}$, and $I_t$) at a time, as the historical information of the bounding boxes and optical flow features is propagated through the hidden states of the two GRUs since \textit{the first appearance} of the object. Specifically, the bounding box $X_t$ and the flattened optical flow features $f_t^\text{obj}$ are processed at time $t+1$ to update the hidden state of the two GRUs:
\begin{equation}
\begin{aligned}
h_t^\text{box} &= \text{GRU} \left(h_{t - 1}^\text{box}, \phi_\text{encoder}^\text{box} (X_t) \right), \\
h_t^\text{flow} &= \text{GRU} \left( h_{t - 1}^\text{flow}, \phi_\text{encoder}^\text{flow} (f_t^\text{obj}) \right),
\end{aligned}
\end{equation}
where $\phi_\text{encoder}^\text{box}$ and $\phi_\text{encoder}^\text{flow}$ are two non-linear functions. The image pipeline uses a MobileNetV2 backbone~\citep{sandler2018mobilenetv2} pretrained on semantic segmentation task in Cityscapes~\citep{cordts2016cityscapes}. We construct the feature generator for images by truncating the MobileNetV2 right after the last bottleneck block and appending a few additional convolution and fully connected layers, summarized as a non-linear function $\phi_\text{encoder}^\text{image}$:
\begin{equation}
e_t^\text{image} = \phi_\text{encoder}^\text{image} \left( \text{MobileNetV2} (I_t) \right),
\end{equation}
where $e_t^\text{image}$ is the image embedding at time $t$.

Considering a relatively long prediction horizon, we initialize the hidden state of the decoder GRU with a fusion of the latest information \textit{at time} $t$ from all three encoders:
\begin{equation}
h_{0|t}^\text{decoder} = \phi_\text{encoder}^\text{fusion} \left( \left[\text{AvgPool}(h_t^\text{box}, h_t^\text{flow}), e_t^\text{image}\right] \right),
\end{equation}
where $\phi_\text{encoder}^\text{fusion}$ is another non-linear function and AvgPool is the element-wise average of two vectors, following~\cite{yao2022dota,yao2019unsupervised}. Based on the information up until time $t$, the model recurrently outputs the parameters $P_{t+k|t}, 1 \le k \le \delta$ for synthesizing the predicted bounding box from the anchor box by following~(\ref{eq:new-box-from-anchor-box}) within the prediction horizon. The anchor box is set to be the last bounding box $X_t$, and thus the first input to the decoder GRU $P_{t|t}$ is a zero vector. More specifically, the parameters for box transformations are generated in an autoregressive manner:
\begin{equation}
\begin{aligned}
h_{k|t}^\text{decoder} &= \text{GRU} \left( h_{k - 1|t}^\text{decoder}, P_{t + k - 1|t} \right), \\
P_{t + k|t} &= \psi_\text{decoder}^\text{box} (h_{k|t}^\text{decoder}),
\end{aligned}
\end{equation}
where $h_{k|t}^\text{decoder}$ is the $k$-th hidden state of the decoder GRU that is initialized with the information up to time $t$, $\psi_\text{decoder}^\text{box}$ is a non-linear funciton, and $k \in (1, \cdots, \delta)$. We train the behavior expert with the mean squared error (MSE) between the predicted and target future bounding boxes.

For each object, the behavior expert predicts bounding boxes of the next $\delta$ time steps until the tracking is lost. Therefore, at time $t$, each object has $\delta$ bounding boxes predicted from time $t - \delta$ to $t - 1$, respectively\endnote{Within the first $\delta$ time steps since the object is detected (i.e., when $1 \le t \le \delta$), there will be only $t - 1$ bounding boxes predicted from the previous time steps at time $t$. After the tracking of the object is lost, there will be only $\delta - \Delta t_\text{lost}$ bounding boxes predicted from the previous time steps at time $t$, where $\Delta t_\text{lost}$ is the number of elapsed time steps since the lost of tracking.}. It has been shown that monitoring the \textit{prediction consistency} delivers superior AD performance than monitoring the prediction accuracy. In particular, at time $t$, the \textit{standard deviation} of the $\delta$ predicted bounding boxes from the previous time steps is computed and averaged over all tracked objects as the anomaly score~\citep{yao2022dota}:
\begin{equation}
\label{eq:fol-anomaly-score}
s_t^\text{std} = \frac{1}{4N} \sum_{i=1}^N \sum_{\hat{y} \in \{\hat{c}^x,\hat{c}^y,\hat{w},\hat{h}\}} \text{STD} ((\hat{y}_{t|t - \delta}^i, \cdots, \hat{y}_{t|t - 1}^i)),
\end{equation}
where $N$ is the number of tracked objects at time $t$. A low STD indicates that the behavior of the object follows normal patterns and thus the predictions are stable, while a high STD suggests anomalous behaviors (e.g., the predictions made before and after a collision for a specific future time step are highly inconsistent).

\begin{figure}[t]
  \centering
  \begin{subfigure}[b]{0.09\linewidth}
    \captionsetup{justification=centering}
    \includegraphics[width=\linewidth]{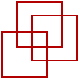}
    \caption{}
    \label{subfig:height-scaling-eg-small}
  \end{subfigure} \hspace{5mm}
  \begin{subfigure}[b]{0.3\linewidth}
    \captionsetup{justification=centering}
    \includegraphics[width=\linewidth]{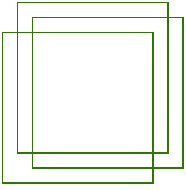}
    \caption{}
    \label{subfig:height-scaling-eg-large}
  \end{subfigure}
  \caption{\textbf{An illustrative example showing the limitation of the original anomaly score.} The bounding box predictions for a small object and a large object are shown in (a) and (b), respectively. The two cases result in identical STD values even though the prediction consistency is perceptually different.}
  \label{fig:height-scaling-eg}
\end{figure}

In this paper, we make a simple yet important modification to~(\ref{eq:fol-anomaly-score}) by scaling the STD value with the height of the bounding boxes to further improve the AD performance. It is empirically observed that the prediction consistency is positively correlated with the distance from the ego car to the object. In other words, future locations of a more distant and smaller object can be predicted more stably than a closer and larger object, possibly due to the fact that adjacent objects often exhibit large motions because of perspective projection. Formally, let $\hat{y}$ be one of the elements of a predicted bounding box vector $[\hat{c}^x, \hat{c}^y, \hat{w}, \hat{h}]$ and $y$ be the corresponding ground truth element. For simplicity, we assume that a prediction \textit{in normal cases} is a sample of a random variable with a Gaussian distribution, whose mean is the ground truth value and variance is \textit{proportional} to the power of the height of the predicted bounding box, and that the prediction error is agnostic to the time step. Thus, by omitting the subscript of time steps, we arrive at $\hat{y} \sim \mathcal{N}(y, \alpha \hat{h}^\beta)$, where $\alpha$ and $\beta$ are two positive numbers. We further assume that $\beta=2$ for simplicity. As a result, the STD of $\hat{y}$'s within the prediction horizon is $\sqrt{\alpha} \hat{h}$, which is then used to compute the anomaly score. However, such a score can be misleading, as whether or not an object's behavior is anomalous should not depend on the height of the object. Therefore, we simply scale the STD value by $\hat{h}$ to resolve such an issue.

Figure~\ref{fig:height-scaling-eg} shows the benefit of such a scaled loss from another perspective. In this illustrative example, we keep the size of the predicted bounding box unchanged \textit{within each case} and shift the center of the bounding box by the same amount \textit{across the two cases}. As a result, the anomaly scores computed for the small and the large object are identical according to~(\ref{eq:fol-anomaly-score}). However, it is apparent that the predictions in Figure~\ref{subfig:height-scaling-eg-small} are less consistent than those in Figure~\ref{subfig:height-scaling-eg-large}, and an anomaly is more likely to be present in~\ref{subfig:height-scaling-eg-small}. By scaling the STD values with the height of the predicted bounding box, we obtain a higher anomaly score for~\ref{subfig:height-scaling-eg-small} than~\ref{subfig:height-scaling-eg-large}, which meets our expectations. The final anomaly score from the behavior expert is computed as:
\begin{equation}
s_t^\text{beh} = \frac{1}{4N} \sum_{i=1}^N \lambda_{\hat{h}}^i \sum_{\hat{y} \in \hat{Y}} \text{STD} ((\hat{y}_{t|t - \delta}^i, \cdots, \hat{y}_{t|t - 1}^i)),
\end{equation}
where $\hat{Y} = \{\hat{c}^x,\hat{c}^y,\hat{w},\hat{h}\}$ and $\lambda_{\hat{h}}^i$ is the average height of the predicted bounding boxes for the $i$-th object at time $t$:
\begin{equation}
\lambda_{\hat{h}}^i = \frac{1}{\delta} \sum_{k=1}^\delta \hat{h}_{t|t - k}^i.
\end{equation}
Similar to the interaction expert, we also apply a digital low-pass filter on top of $s_t^\text{beh}$ for smoothness over time.

%% file: Sections/05-Ensemble.tex
\section{Expert ensemble}
The scene, interaction, and behavior expert are designed for different types of anomalies, as identified in Figure~\ref{fig:anomaly-patterns}. An efficient ensemble method is required to build a robust AD system as a driving scenario can be accompanied with any kinds of anomalies. In this work, we adopt a result-level fusion by combining the anomaly scores from all the experts to generate a final anomaly score at time $t$:
\begin{equation}
s_t = e( s_t^\text{ffp}, s_t^\text{str}, s_t^\text{int}, s_t^\text{beh} ),
\end{equation}
where $e$ is the ensemble function. We now present two necessary steps towards the final score during evaluation.

\subsection{Expert score normalization}
The anomaly scores produced by different experts can differ in orders of magnitude due to the completely different mechanisms of score generation. To balance the effect of each score on the final anomaly score, we perform a score normalization step for each expert during evaluation:
\begin{equation}
\label{eq:score-normalization}
\bar{s}_t^{\,\text{exp}} = \frac{s_t^\text{exp} - \mu}{\sigma},
\end{equation}
where $\bar{s}_t^{\,\text{exp}}$ is the normalized expert score, $s_t^\text{exp} \in \{s_t^\text{ffp}, s_t^\text{str}, s_t^\text{int}, s_t^\text{beh}\}$, and $\mu$ and $\sigma$ are the mean and standard deviation of the expert score on the \textit{training data} without any anomalies, respectively. The normalization~(\ref{eq:score-normalization}) differs slightly from the z-score normalization in that $\mu$ and $\sigma$ are computed on the training dataset while $s_t^{\text{exp}}$ comes from a \textit{mutually different} evaluation dataset, which includes both normal and anomalous data. As a result, only $\bar{s}_t^\text{\,exp}$'s obtained from the normal data are expected to have a mean around $0$ and a standard deviation around $1$, while those obtained from the anomalous scenarios are not.

The amount of training data may be limited. Therefore, to estimate a more realistic mean and standard deviation in~(\ref{eq:score-normalization}), we employ a kernel density estimation~\citep{wkeglarczyk2018kernel} to fit the probability density function (pdf) for the normal expert scores. We use a Gaussian kernel and apply the transformation trick~\citep{shalizi2013advanced} to make sure that the estimated pdfs have support on $[0, +\infty]$, as all the expert scores are guaranteed to be positive. The mean and the standard deviation are then computed from the numerical integration on the estimated pdf. We note that only the information derived from the training data is required and thus the computation can be done before model deployment.

In the meantime, we set a threshold for AD for each expert. The threshold is determined by $\tau^\text{exp} = \mathbf{U} (\alpha)$, which is the upper $\alpha$-quantile of the estimated pdf and $\alpha$ is the confidence level, similar to~\citep{feng2022unsupervised}. When evaluating each expert \textit{individually}, we declare an anomaly whenever the unnormalized expert score $s_t^\text{exp}$ is larger than the threshold $\tau^\text{exp}$.

\subsection{Kalman filter based score fusion}
After the score normalization, all the expert scores are brought to a similar range and are ready for the final fusion. However, although informative, the outputs from the experts can be noisy in terms of conveying how likely a type of anomaly is present. In this work, we view each normalized expert score $\bar{s}_t^{\,\text{exp}}$ as a \textit{noise-corrupted} observation of the system state $x_t^\text{exp}$, which reflects the ground truth likelihood of encountering a specific type of anomaly. As mentioned in Section~\ref{sec:overview}, it is possible that each point in time contains more than one anomaly. Therefore, we define the final anomaly score $s_t$ as the addition of the underlying system states for each type of anomaly. Our state vector then becomes $\mathbf{x}_t = [x_t^\text{ffp}, x_t^\text{str}, x_t^\text{int}, x_t^\text{beh}, s_t]^\top \in \mathbb{R}^5$, and the system model has the form of:
\begin{equation}
\label{eq:system-dynamics}
\begin{aligned}
\mathbf{x}_{t+1}
&=
Ax_t + \mathbf{w}_t \\
\mathbf{y}_t
&=
H \mathbf{x}_t + \mathbf{v}_t,
\end{aligned}
\end{equation}
where the observation $\mathbf{y}_t = [\bar{s}_t^\text{\,ffp}, \bar{s}_t^\text{\,str}, \bar{s}_t^\text{\,int}, \bar{s}_t^\text{\,beh}]^\top$ and the state-transition matrix $A$ and the observation matrix $H$ are set respectively as:
\begin{equation}
A
=
\begin{bmatrix}
1 & 0 & 0 & 0 & 0 \\
0 & 1 & 0 & 0 & 0 \\
0 & 0 & 1 & 0 & 0 \\
0 & 0 & 0 & 1 & 0 \\
\frac{1}{4} & \frac{1}{4} & \frac{1}{4} & \frac{1}{4} & 0
\end{bmatrix}, \;
H
=
\begin{bmatrix}
1 & 0 & 0 & 0 & 0 \\
0 & 1 & 0 & 0 & 0 \\
0 & 0 & 1 & 0 & 0 \\
0 & 0 & 0 & 1 & 0
\end{bmatrix}.
\end{equation}
The process noise $\mathbf{w}_t$ and the observation noise $\mathbf{v}_t$ are drawn from two zero mean normal distributions with covariance $Q$ and $R$, respectively. Although the coefficients for the weighted sum in the last row of $A$ can be different, we use identical values for simplicity.

To estimate the final anomaly score, we employ Kalman filter, which has been widely used for estimating the internal state of a system in various application domains~\citep{bewley2016simple,wojke2017simple,sun2021idol}. The state dynamics of our Kalman filter follows the standard framework, such as the one in~\cite{patel2013moving}, and is omitted here for brevity. We set the covariance of the observation noise $R$ to be an identity matrix as a result of the expert score normalization. Using $1$-based indexing, the initial a posteriori estimate covariance matrix $P_{1|1}$ in Kalman filter and the covariance of the process noise $Q$ are both set to be a diagonal matrix filled with $0.1$ for simplicity. The initial guess of the state vector is:
\begin{equation}
\label{eq:kf-initialization}
\begin{aligned}
\hat{\mathbf{x}}_1
&=
[\bar{s}_1^\text{\,ffp}, \bar{s}_1^\text{\,str}, \bar{s}_1^\text{\,int}, \bar{s}_1^\text{\,beh}, s_1]^\top, \\
s_1
&=
0.25(\bar{s}_1^\text{\,ffp} + \bar{s}_1^\text{\,str} + \bar{s}_1^\text{\,int} + \bar{s}_1^\text{\,beh})
\end{aligned}
\end{equation}
The final anomaly score at time $t$ is retrieved as the last element of the state estimate $\hat{\mathbf{x}}_t$ from Kalman filter.

To obtain a threshold for AD for the ensemble, we normalize each $\tau^\text{exp}$ similar to~(\ref{eq:score-normalization}) and perform a weighted sum using the same weights as those in the score fusion: $\tau = 0.25(\bar{\tau}^\text{ffp} + \bar{\tau}^\text{str} + \bar{\tau}^\text{int} + \bar{\tau}^\text{beh})$.

%% file: Sections/06-Experiments.tex
\section{Experimental results}
\label{sec:experiments}

In our experiments, we evaluate the anomaly detection performance of Xen on the largest on-road video anomaly dataset, named the Detection of Traffic Anomaly (DoTA) dataset~\citep{yao2022dota}. DoTA is comprised of diverse video clips collected from dashboard cameras in different areas (e.g., East Asia, North America, Europe) under different weather (e.g., sunny, cloudy, raining, snowing) and lighting conditions (day and night). Each video in the dataset is recorded at $10$ fps and contains one anomalous event. Due to different video sources, camera intrinsics and poses are different across videos.

We use the same training and test split of DoTA as those in~\cite{yao2022dota} and further augment the test split with $88$ videos, which are provided in the dataset but unlabeled, by following the annotation principles in the original paper. The final training set consists of 3275 videos and the test set contains 1490 videos. Each video is annotated with anomaly start and end time, based on which a frame is labeled as $0$ or $1$ for a normal or anomalous time step, respectively, for evaluation purposes only. Notably, the anomaly start time in DoTA is defined as the time when the anomaly is inevitable, which is often prior to an actual accident (e.g., a car crash). Therefore, the evaluation reflects the model performance of early anomaly detection.

There exist $9$ categories of on-road anomaly in DoTA, and each category is further split into ego involved and non-ego involved cases, resulting in $18$ categories in total. For quick reference, we list the $9$ ego involved categories in Table~\ref{table:anomaly-category-DoTA}. To represent non-ego involved counterparts, we append a letter of ``N" to each label in Table~\ref{table:anomaly-category-DoTA}. For example, ST-N stands for non-ego involved collisions with another vehicle that starts, stops, or is stationary. We refer to~\cite{yao2022dota} for data samples in DoTA. Based on the analysis on common anomaly patterns in Section~\ref{sec:overview}, we classify all the anomaly categories in Table~\ref{table:anomaly-category-DoTA} as \textit{ego involved anomalies}, VO-N, OO-N, UK-N as \textit{non-ego involved individual anomalies}, and the rest $6$ categories as \textit{non-ego involved interactive anomalies}. Unsupervised AD models can only be trained with normal data, so we use the frames before the anomaly start time in each video in the training set for training. During evaluation, in contrast to the DoTA paper where the videos under unknown categories or without objects were ignored~\citep{yao2022dota}, we evaluate on all test videos as we believe that an anomaly detector needs to be robust in any driving scenarios.

\begin{table}[t]
  \begin{center}
    \caption{On-road anomaly category in DoTA~\citep{yao2022dota}.}
    \label{table:anomaly-category-DoTA}
    \resizebox{\linewidth}{!}{%
    \begin{tabular}{ l | l }
      \toprule
      Label & Anomaly Category Description \\
      \midrule
      ST & \makecell[lt]{Collision with another vehicle that starts, stops, \\ or is stationary} \\
      AH & Collision with another vehicle moving ahead or waiting \\
      LA & \makecell[lt]{Collision with another vehicle moving laterally in \\ the same direction} \\
      OC & Collision with another oncoming vehicle \\
      TC & \makecell[lt]{Collision with another vehicle that turns into or \\ crosses a road} \\
      VP & Collision between vehicle and pedestrian \\
      VO & Collision with an obstacle in the roadway \\
      OO & Out-of-control and leaving the roadway to the left or right \\
      UK & Unknown \\
      \bottomrule
    \end{tabular}}
  \end{center}
\end{table}

To realize Xen, we train the scene expert (including FFP and STR), the interaction expert, and the behavior expert individually. An Adam optimizer~\citep{kingma2014adam} without weight decay is used for all the experts. Common training hyperparameters are listed in Table~\ref{table:training-hyperparameters}. The model-specific implementation details are described as follows.

\paragraph{Scene expert:} The disparity values are clipped at $2000$ before normalization. The weights in the training objective~(\ref{eq:str-objective}) are specified as $\lambda_\text{d}=100$ and $\lambda_\text{tv}=0.1$.

\paragraph{Interaction expert:} The history horizon $T_\text{int}$ is set to be $3$. The embedding function $\phi_\text{encoder}$ is constructed by $2$ fully connected (FC) layers with $(32, 64)$ hidden units. The non-linear function $\phi_\text{z}$ is implemented as one FC layer with $4$ hidden units. Both the encoder and decoder GRU use a hidden size of $128$. In the decoder, the embedding function $\phi_\text{decoder}$ has the same structure as $\phi_\text{encoder}$, and the function $\psi_\text{decoder}$ is constructed by $2$ FC layers with $(64, 8)$ hidden units. ReLU activation functions are applied after each FC layer except for the last layer in $\psi_\text{decoder}$. The capacity of the priority queue $N_\text{max}$ is set to be $20$.

\paragraph{Behavior expert:} The prediction horizon $\delta$ is set to be $10$. The function $\phi_\text{encoder}^\text{box}$ and $\phi_\text{encoder}^\text{flow}$ have the same structure of $2$ FC layers with $(512, 64)$ hidden units. The embedding function $\phi_\text{encoder}^\text{image}$ is constructed by $2$ convolution layers with filter number $(160, 32)$ and filter size $3 \times 3$, a flattening operation, and one FC layer with $512$ hidden units. Each convolution layer is followed by a $2 \times 2$ max pooling layer. We implement $\phi_\text{encoder}^\text{fusion}$ with $2$ FC layers with $(512, 512)$ hidden units. All the GRUs have a hidden size of $512$. In the decoder, the non-linear function $\psi_\text{decoder}$ is constructed by $2$ FC layers with $(32, 4)$ hidden units. ReLU activation functions are applied after each max pooling layer and FC layer except for the last layer in $\phi_\text{encoder}^\text{fusion}$ and $\psi_\text{decoder}^\text{box}$.

Both the interaction and behavior expert use a low-pass Butterworth filter of order $2$ and cut-off frequency $0.2$ Hz.

\begin{table}[t]
  \begin{center}
    \caption{Determination of training hyperparameters. The learning rate is kept constant during training for each model.}
    \label{table:training-hyperparameters}
    \resizebox{\linewidth}{!}{%
    \begin{tabular}{ l | c  c  c  c }
      \toprule
      Hyperparameter & FFP & STR & Interaction expert & Behavior expert \\
      \midrule
      Batch size & 4 & 4 & 64 & 16 \\
      Learning rate & 0.0002 & 0.0002 & 0.0002 & 0.0005 \\
      \bottomrule
    \end{tabular}}
  \end{center}
\end{table}

\subsection{Baselines and metrics}
We evaluate the performance of the proposed method on the test set, along with the following baseline methods:
\begin{itemize}
\item
\textit{Conv-AE}~\citep{hasan2016learning}: An AE-based model which encodes temporally stacked RGB images with convolution layers and decodes the resulting features with deconvolution layers to reconstruct the input. The MSE over pixels in a frame is computed as the anomaly score.
\item
\textit{AnoPred}~\citep{liu2018future}: A unet-based model which takes four preceding RGB frames as input to directly predict a future frame. The negative PSNR of the predicted frame is used as the anomaly score.
\item
\textit{FOL}~\citep{yao2019unsupervised}: A recurrent encoder-decoder network which sequentially processes the past ego motions, object's bounding boxes, and optical flow features to predict future locations of the object over a short horizon. Among the three strategies of computing anomaly scores proposed in the original paper, we use the best-performing prediction consistency~(\ref{eq:fol-anomaly-score}) as our baseline.
\item
\textit{FOL-Ensemble}~\citep{yao2022dota}: An ensemble method which achieves the state-of-the-art results on the DoTA dataset. The object anomaly score, generated by FOL, is first mapped to per-pixel scores by putting a Gaussian function at the center of each object. Such a pseudo anomaly score map is then averaged with the error map generated by AnoPred to produce the fused anomaly score map. The averaged score over all the pixels in the fused score map serves as the final anomaly score.
\end{itemize}

All the baselines are adapted to DoTA based on the corresponding released code and original paper. We also perform an expert-level ablation study by sequentially removing the experts from Xen to investigate the necessity of each module. In the rest of this section, we use \textbf{Xen-S} to denote the ensemble of FFP and STR in the scene expert, \textbf{Xen-SI} the ensemble of the scene and interaction expert, and \textbf{Xen-SIB} the ensemble of all the three experts. The Kalman filter design remains unchanged for the ablated versions of Xen-SIB with only necessary modifications to account for the reduced dimensionality of the system. For a fair comparison, we use the same optical flow estimation network and object detection and tracking network to generate data for all the benchmarked methods if needed. All the models are trained on the same dataset. During evaluation, at the beginning of a video when a method is not able to produce an anomaly score due to insufficient observations as input, we simply assign a score assuming that the frame is normal.

In Xen, the start time of Kalman filtering plays an important role in final results. By default, we start the filtering only when all the experts start to produce anomaly scores, otherwise we keep applying the initialization step~(\ref{eq:kf-initialization}) by changing the subscript to the corresponding time step. An alternative way is to start the filtering immediately as a video starts. In other words, the initialization~(\ref{eq:kf-initialization}) is applied only at time $1$ and the assigned scores serve as the measurements of Kalman filter at the first few frames of a video. We mark the variants using the second filtering mechanism with $^\star$ in the rest of this paper.

Quantitatively, we compare different methods using the following two metrics:
\begin{itemize}
\item
\textit{F1-score}: A comprehensive threshold-dependent index considering precision and recall, which can be written as $2 \cdot \text{precision} \cdot \text{recall} / (\text{precision} + \text{recall})$.
\item
\textit{AUC}: A threshold-independent index indicating the area under the receiver operating characteristics curve. AUC describes the ability to distinguish between positive and negative samples for anomaly detection models.
\end{itemize}

\begin{figure}[t]
  \centering
  \begin{subfigure}[b]{0.49\linewidth}
    \captionsetup{justification=centering}
    \includegraphics[width=\linewidth]{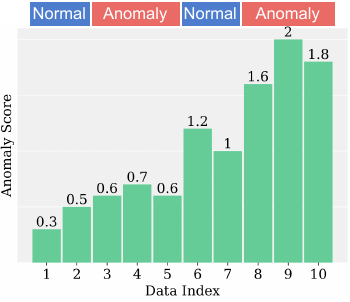}
    \caption{Without normalization}
  \end{subfigure}
  \begin{subfigure}[b]{0.49\linewidth}
    \captionsetup{justification=centering}
    \includegraphics[width=\linewidth]{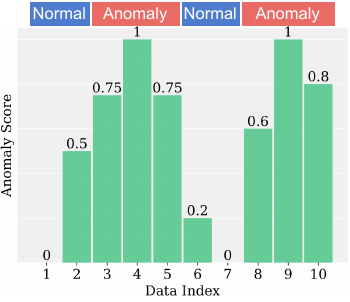}
    \caption{With normalization}
  \end{subfigure}
  \caption{\textbf{An illustrative example showing that the video-wise min-max normalization distorts the actual model performance.} The results from two videos are concatenated. The ground truth label is annotated on top. Frame-level anomaly score is shown above each bar.}
  \label{fig:metrics-eg}
\end{figure}

Following common practice in video AD literature, we concatenate the anomaly score vectors of all the videos before computing the metrics rather than average the metrics of each video. However, prior works often apply a video-wise min-max normalization on scores before the concatenation~\citep{hasan2016learning,liu2018future,liu2021hybrid,yao2022dota}. Such a normalization step can lead to overoptimistic performance, especially when each video contains an anomaly. For example, consider a case where video A of length $5$ has a ground truth label of $(0, 0, 1, 1, 1)$ and predicted anomaly scores of $(0.3, 0.5, 0.6, 0.7, 0.6)$ and video B has the same ground truth but different anomaly scores of $(1.2, 1.0, 1.6, 2.0, 1.8)$. The concatenated scores with and without video-wise normalization is shown in Figure~\ref{fig:metrics-eg}. It is apparent that such an anomaly detector is not perfect since the predicted anomaly scores of the last three \textit{anomalous} frames in video A are lower than those of the first two \textit{normal} frames in video B. With video-wise normalization, however, the results falsely suggest that the anomaly detector is perfect with an AUC of $1$. From another perspective, the min-max normalization implicitly provides additional information that each video contains at least one anomaly as the maximum anomaly score is guaranteed to be $1$ in each video. Furthermore, the normalization is simply impossible in online AD. Therefore, we compute the metrics without video-wise min-max normalization for more realistic evaluation in this work.

\begin{table}[t]
  \begin{center}
    \caption{Comparison of anomaly detection performance with different methods on DoTA. The variants of Xen without $^\star$ start Kalman filtering only when all the experts start to produce anomaly scores, while those with $^\star$ start filtering since the video start time.}
    \label{table:overall-results}
    \resizebox{\linewidth}{!}{%
    \begin{tabular}{  l  l  c  c  }
      \toprule
      Method & Input & AUC & F1-score \\
      \midrule
      Conv-AE & RGB & $58.7$ & $50.8$ \\
      AnoPred & RGB & $67.9$ & $56.0$ \\
      FOL & Flow + Box + Ego & $65.0$ & $53.7$ \\
      FOL-Ensemble & RGB + Flow + Box + Ego & $68.4$ & $56.2$ \\
      \midrule
      Xen-S & RGB + Flow + Depth & $69.1$ & $56.4$ \\
      Xen-SI & RGB + Flow + Depth + Box & $72.5$ & $58.2$ \\
      Xen-SIB & RGB + Flow + Depth + Box & $\mathbf{74.4}$ & $\mathbf{60.0}$ \\
      \midrule
      Xen-S$^\star$ & RGB + Flow + Depth & $71.4$ & $57.6$ \\
      Xen-SI$^\star$ & RGB + Flow + Depth + Box & $73.9$ & $59.4$ \\
      Xen-SIB$^\star$ & RGB + Flow + Depth + Box & $\mathbf{75.1}$ & $\mathbf{60.5}$ \\
      \bottomrule
    \end{tabular}}
  \end{center}
\end{table}

\subsection{Overall results}
The overall results are presented in Table~\ref{table:overall-results}. As shown, Xen-SIB achieves the best AUC and highest F1-score with a large margin over the four baselines. Although we were able to reproduce the performance of FOL in the original paper~\citep{yao2022dota} with $0.1$ difference in AUC, the AD metrics drop significantly after evaluating on all test videos and removing the video-wise min-max normalization. We argue that this is partially due to the fact that object-centric methods can completely fail in \textit{ego involved anomalies} where no traffic participants are detected within the field of view. FOL-Ensemble alleviates such a problem by fusing the scores from the frame-level method AnoPred and the object-centric method FOL. In fact, all the ensemble methods (e.g., Xen-SI and Xen-SIB), which combine frame-level and object-level features for AD, show superior AUC and F1-score compared to their individual component (e.g., Xen-S) in the ensemble. With a closer observation, we found that the STR alone in the scene expert produces an AUC of $64.1$, which is much higher than that from Conv-AE. Considering that STR and Conv-AE are both reconstruction-based method and have similar architecture, such a result indicates that the spatiotemporal representation (i.e., optical flow and disparity) of a scene is more effective as inputs for AD than the simple RGB representation. By combining the reconstruction-based STR and the prediction-based FFP through Kalman filter, our Xen-S can already achieve reasonable improvement over the baselines.

To our knowledge, Xen-SIB is the first that is built upon a holistic analysis on anomaly patterns. Our ablation study shows that the detection of each edge in Figure~\ref{fig:anomaly-patterns} contributes to the overall performance of Xen-SIB. Specifically, Xen-SI achieves higher AUC and F1-score than Xen-S by including the edge of \textit{non-ego involved interactive anomalies} so that anomalous interactions between other road participants can be monitored and detected. Xen-SIB, which encompasses all the edges in Figure~\ref{fig:anomaly-patterns}, shows the best AD performance among all the ablated versions of Xen-SIB due to the ability to model the most diverse normal patterns in driving scenarios. Notably, compared to applying Kalman filter after all the experts start to produce anomaly scores, filtering immediately since time $1$ consistently performs better in all the ablations. Such an improvement is expected. On the one hand, videos often start with normal scenarios. On the other hand, we assign scores assuming that the frame is normal for an expert at the beginning of a video when the observations are insufficient to make an inference. The effects of these assigned scores are then propagated through the system dynamics~(\ref{eq:system-dynamics}), biasing the immediately subsequent anomaly scores to be low. Although making use of assigned scores, such a filtering mechanism is indeed applicable in the real world. However, the performance gap between the two filtering techniques will be smaller with the increase of the video length.

\begin{table}[t]
  \begin{center}
    \caption{Comparison of anomaly detection performance with different methods in ego involved and non-ego involved cases. The variants of Xen without $^\star$ start Kalman filtering only when all the experts start to produce anomaly scores, while those with $^\star$ start filtering since the video start time. In general, frame-level methods (i.e., Conv-AE, AnoPred, and Xen-S) perform well in ego involved cases, object-centric methods (i.e., FOL) excel in non-ego involved cases, and ensemble methods take advantage of both types of methods.}
    \label{table:ego-non-ego-results}
    \resizebox{\linewidth}{!}{%
    \begin{tabular}{ l  c  c  c  c  c }
      \toprule
      \multirow{2}{*}{Method} & \multicolumn{2}{c}{Ego involved} & & \multicolumn{2}{c}{Non-ego involved} \\
                              & AUC & F1-score & & AUC & F1-score \\
      \midrule
      Conv-AE & $63.9$ & $51.9$ & & $51.2$ & $49.5$ \\
      AnoPred & $73.0$ & $58.7$ & & $60.8$ & $52.6$ \\
      FOL & $64.8$ & $53.2$ & & $65.6$ & $54.3$ \\
      FOL-Ensemble & $73.4$ & $59.2$ & & $61.3$ & $52.4$ \\
      \midrule
      Xen-S & $74.4$ & $58.9$ & & $62.0$ & $53.2$ \\
      Xen-SI & $76.0$ & $60.7$ & & $67.7$ & $54.9$ \\
      Xen-SIB & $\mathbf{77.3}$ & $\mathbf{62.3}$ & & $\mathbf{70.6}$ & $\mathbf{57.1}$ \\
      \midrule
      Xen-S$^\star$ & $76.7$ & $60.1$ & & $64.5$ & $54.3$ \\
      Xen-SI$^\star$ & $77.4$ & $62.2$ & & $69.1$ & $55.9$ \\
      Xen-SIB$^\star$ & $\mathbf{77.9}$ & $\mathbf{62.9}$ & & $\mathbf{71.3}$ & $\mathbf{57.4}$ \\
      \bottomrule
    \end{tabular}}
  \end{center}
\end{table}

\begin{table*}[t]
  \begin{center}
    \caption{Comparison of anomaly detection performance with different methods on each type of anomaly. Conv-AE, Xen-S, and the variants of Xen using immediate filtering since the video start time are omitted for brevity.}
    \label{table:fine-grained-results}
    \resizebox{\linewidth}{!}{%
    \begin{tabular}{ l  c  c  c  c  c  c  c  c  c  c  c  c  c  c  c  c  c  c  c  c }
      \toprule
      & \multicolumn{9}{c}{Ego} & & \multicolumn{6}{c}{Non-ego interactive} & & \multicolumn{3}{c}{Non-ego individual} \\
      \cmidrule{2-10}
      \cmidrule{12-17}
      \cmidrule{19-21}
      & ST & AH & LA & OC & TC & VP & VO & OO & UK & & ST-N & AH-N & LA-N & OC-N & TC-N & VP-N & & VO-N & OO-N & UK-N \\
      \midrule
      Method & \multicolumn{20}{c}{AUC} \\
      \midrule
      AnoPred & $72.3$ & $75.7$ & $74.7$ & $68.7$ & $73.6$ & $70.2$ & $78.4$ & $72.0$ & $64.9$ & & $63.6$ & $58.4$ & $59.6$ & $63.9$ & $62.6$ & $64.2$ & & $57.9$ & $59.0$ & $60.8$ \\
      FOL & $72.7$ & $73.4$ & $66.5$ & $\mathbf{76.6}$ & $72.4$ & $67.1$ & $54.4$ & $44.1$ & $56.5$ & & $\mathbf{75.2}$ & $\mathbf{68.2}$ & $62.8$ & $66.8$ & $68.4$ & $74.8$ & & $57.3$ & $64.8$ & $60.9$ \\
      FOL-Ensemble & $73.0$ & $76.4$ & $75.3$ & $69.3$ & $74.3$ & $70.1$ & $78.9$ & $71.9$ & $65.1$ & & $65.3$ & $59.0$ & $59.8$ & $64.4$ & $63.4$ & $65.6$ & & $58.3$ & $59.3$ & $61.5$ \\
      Xen-SI (ours) & $71.8$ & $79.1$ & $\mathbf{79.4}$ & $71.0$ & $76.8$ & $69.2$ & $79.5$ & $\mathbf{74.8}$ & $70.9$ & & $70.9$ & $65.9$ & $65.0$ & $72.1$ & $70.7$ & $73.7$ & & $60.0$ & $66.6$ & $67.4$ \\
      Xen-SIB (ours) & $\mathbf{74.5}$ & $\mathbf{80.2}$ & $\mathbf{79.5}$ & $75.7$ & $\mathbf{79.2}$ & $\mathbf{72.7}$ & $\mathbf{80.0}$ & $74.1$ & $\mathbf{71.5}$ & & $70.5$ & $67.9$ & $\mathbf{66.8}$ & $\mathbf{75.5}$ & $\mathbf{72.3}$ & $\mathbf{77.1}$ & & $\mathbf{60.2}$ & $\mathbf{72.3}$ & $\mathbf{71.2}$ \\
      \midrule
      Method & \multicolumn{20}{c}{F1-score} \\
      \midrule
      AnoPred & $50.0$ & $56.9$ & $62.2$ & $52.4$ & $56.5$ & $50.4$ & $56.1$ & $68.1$ & $57.4$ & & $38.9$ & $51.3$ & $55.7$ & $44.3$ & $49.0$ & $46.7$ & & $50.9$ & $57.7$ & $\mathbf{62.3}$ \\
      FOL & $52.9$ & $55.6$ & $58.0$ & $56.1$ & $55.3$ & $46.6$ & $42.3$ & $35.0$ & $46.6$ & & $\mathbf{47.9}$ & $53.4$ & $54.7$ & $45.8$ & $53.5$ & $53.3$ & & $51.2$ & $58.4$ & $55.6$ \\
      FOL-Ensemble & $51.0$ & $57.7$ & $62.8$ & $52.8$ & $56.9$ & $\mathbf{50.8}$ & $57.9$ & $68.2$ & $57.6$ & & $39.1$ & $51.3$ & $55.5$ & $44.8$ & $49.4$ & $47.3$ & & $\mathbf{51.5}$ & $56.8$ & $61.0$ \\
      Xen-SI (ours) & $52.5$ & $60.5$ & $64.8$ & $53.9$ & $58.2$ & $48.3$ & $\mathbf{60.7}$ & $\mathbf{68.8}$ & $\mathbf{58.5}$ & & $44.0$ & $\mathbf{54.1}$ & $54.7$ & $46.9$ & $54.3$ & $51.7$ & & $47.6$ & $59.2$ & $62.1$ \\
      Xen-SIB (ours) & $\mathbf{54.2}$ & $\mathbf{62.6}$ & $\mathbf{65.5}$ & $\mathbf{57.0}$ & $\mathbf{60.9}$ & $50.3$ & $59.9$ & $67.9$ & $57.2$ & & $47.0$ & $53.4$ & $\mathbf{56.1}$ & $\mathbf{51.6}$ & $\mathbf{55.9}$ & $\mathbf{59.5}$ & & $50.6$ & $\mathbf{62.8}$ & $60.8$ \\
      \bottomrule
    \end{tabular}}
  \end{center}
\end{table*}

\subsection{Per-class results}
Table~\ref{table:ego-non-ego-results} shows the results of different methods broken out into ego involved and non-ego involved cases. In general, non-ego involved anomalies are more difficult to detect than ego involved ones, possibly because non-ego anomalies are often accompanied with small anomalous regions and low camera visibility of objects. By comparing Xen-SI with frame-level methods (e.g., AnoPred and Xen-S), we found that the improvement in non-ego involved cases is much more significant than that in ego involved cases, which is expected since the interaction expert mainly enhances the object-level AD performance. A similar improvement is observed when further including the behavior expert in Xen-SIB, which achieves the best AUC and F1-score in both ego involved and non-ego involved cases. By contrast, the object-centric method FOL performs relatively well in non-ego involved cases but struggles to excel in ego involved cases, resulting in overall inferior performance compared to Xen variants. Despite the fusion of frame-level and object-centric method, FOL-Ensemble fails to provide as large improvements as Xen-SIB does over each component (e.g., AnoPred) in the ensemble, which highlights the importance of effective fusion mechanism for AD in complex driving scenarios. Last but not least, in both ego involved and non-ego involved cases, Xen variants with immediate filtering since time $1$ still consistently outperform those with filtering since the time when all experts start to produce anomaly scores, which agrees with the findings in Table~\ref{table:overall-results}.

More fine-grained results on the AD performance for each type of anomaly are presented in Table~\ref{table:fine-grained-results}. We observe that Xen-SI and Xen-SIB exhibit the highest AUC and F1-score in most columns. FOL demonstrates good AD performance in some of the non-ego involved anomalies (e.g., ST-N and AH-N), but completely fails in several ego involved anomalies (e.g., VO and OO), which is expected as no anomalous road participants can be detected in such cases. Second, by comparing Xen-SI and Xen-SIB, we found that the behavior expert mainly helps improve the detection of \textit{non-ego involved individual anomalies}, which meets our design goal. Meanwhile, \textit{non-ego involved interactive anomalies} are also better detected with the behavior expert in Xen-SIB as interactive anomalies can also contain anomalous behaviors of individual objects. Third, certain classes, especially the non-ego involved ones, are especially challenging for all the anomaly detectors. For example, obstacles in non-ego vehicle-obstacle anomalies (VO-N) are often occluded and the tracking of the impacted vehicle is often lost after the collision, making the abnormal motions invisible to the anomaly detector; and vehicles in non-ego lateral anomalies (LA-N) often move towards each other slowly before the accident, making the anomaly relatively subtle and hard to detect.

\begin{figure*}
  \centering
  \includegraphics[width=0.9\linewidth]{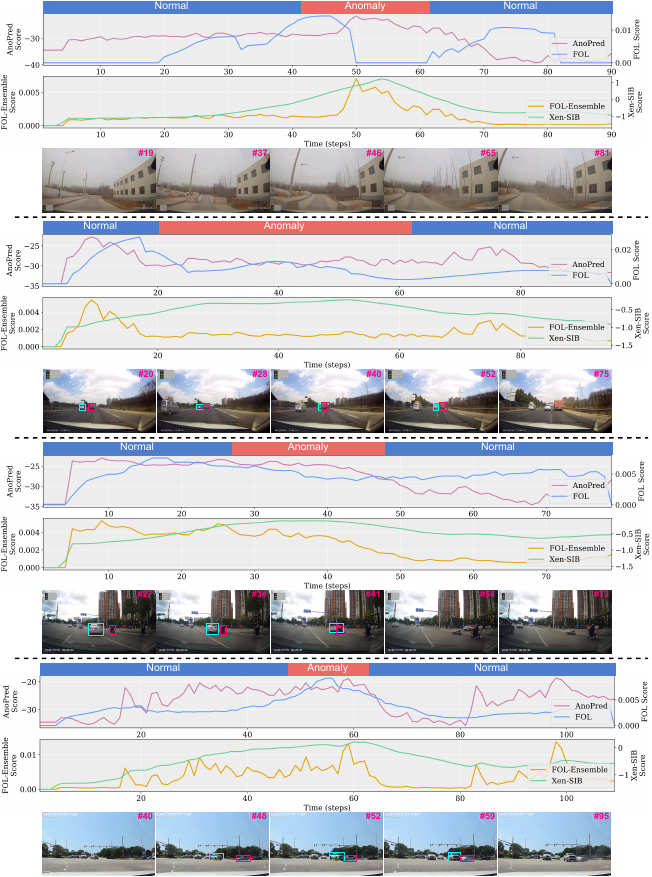}
  \caption{\textbf{Qualitative comparisons of different methods in typical scenarios.} Anomaly scores from different methods lie in different ranges, so each plot contains $2$ scales for different models. The PSNR from AnoPred is negated to serve as the anomaly score. The bounding boxes in sample video frames mark the anomaly participants. The time index of each frame is annotated on the top right corner of the image.}
  \label{fig:qualitative-results}
\end{figure*}

Figure~\ref{fig:qualitative-results} shows the anomaly detection results of different methods in several typical scenarios. In the first example with an ego vehicle-obstacle anomaly (VO), the ego car collides with two roadblocks while moving forward. FOL fails the AD task as no road participants are detected in the anomalous region, and the peaks of the scores are due to the intermittent detection of distant cars in the scene. AnoPred and FOL-Ensemble successfully catch the anomaly after the collision due to the visual inconsistency of the frames but refuse to raise the score at the beginning of the anomaly. By contrast, Xen-SIB is able to produce a high anomaly score earlier than AnoPred and FOL-Ensemble, as the STR in the scene expert can directly detect abnormal scenes (e.g., near-collisions) through spatiotemporal reconstruction. The second example shows a non-ego lateral anomaly (LA-N) between the white car and the red truck. Such an anomaly is challenging for frame-level methods as the anomalous objects only occupy a small region of the scene. Due to this reason, AnoPred fails to increase the score during the anomaly. From the perspective of the ego camera, car motions in this example are not as dramatic as those in a severe proximate accident. As a result, FOL predictions are relatively consistent for each vehicle during the anomaly, leading to miss detection. Thanks to the expert ensemble, Xen-SIB is able to monitor both abnormal individual motions and abnormal relative motions between two road participants and thus is the only method that detects the anomaly successfully. The third example contains a non-ego turning anomaly (TC-N), where the white car collides with a motorcyclist while turning. Xen-SIB generates high scores during the anomaly while the other three methods either produce indistinguishable scores (AnoPred) or result in false alarms (FOL and FOL-Ensemble). The last example demonstrates the importance of object-centric methods through another non-ego turning anomaly (TC-N). Two cars are involved in a side collision at an intersection. AnoPred still fails in such a non-ego involved anomaly. With the fusion of FOL and AnoPred, relatively high anomaly scores appear in the middle of the anomaly using FOL-Ensemble but only last for less than $3$ seconds. By making use of a more effective ensemble of frame-level and object-centric methods, Xen-SIB manages to catch the anomaly over a longer window than the baseline.

\begin{figure*}
  \centering
  \includegraphics[width=0.9\linewidth]{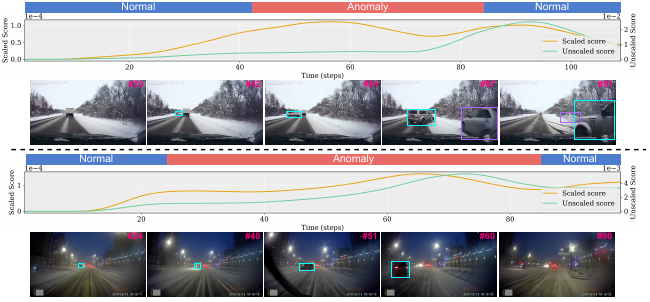}
  \caption{\textbf{Qualitative comparisons of scaled and unscaled anomaly scores from the behavior expert.} The scaled and unscaled scores are generated by the model in the third and fourth row in Table~\ref{table:behavior-expert-ablations}, respectively.}
  \label{fig:behavior-expert-ablation}
\end{figure*}

\subsection{Additional analysis on behavior expert}
Object-centric methods have been shown critical in traffic AD in first-person videos~\citep{yao2019unsupervised,yao2022dota,fang2022traffic}. As one of the most effective object-centric approaches, FOL inspires our work and serves as the backbone of the behavior expert. To further verify the benefits of the modifications we made from FOL for the behavior expert, we perform an additional ablation study. Specifically, we aim to answer the following two questions:
\begin{enumerate*}[label=(\arabic*)]
\item
does each input modality (i.e., bounding boxes, optical flow features, and the RGB image) jointly contribute to the detection of anomalous objects and
\item
does the scaling of scores with bounding box heights benefit the final AD performance?
\end{enumerate*}

\begin{table}[t]
  \begin{center}
    \caption{Ablation study on the behavior expert.}
    \label{table:behavior-expert-ablations}
    \resizebox{\linewidth}{!}{%
    \begin{tabular}{ c  c  c  c  c }
      \toprule
      Bounding box & Optical flow & Image & Scaling & AUC \\
      \midrule
      \checkmark & - & - & \checkmark & $67.0$ \\
      \checkmark & \checkmark & - & \checkmark & $68.2$ \\
      \checkmark & \checkmark & \checkmark & \checkmark & $\mathbf{68.4}$ \\
      \checkmark & \checkmark & \checkmark & - & $66.7$ \\
      \bottomrule
    \end{tabular}}
  \end{center}
\end{table}

We compute the AUC of different ablated versions of the behavior expert on the test set without expert ensemble for a direct comparison. The results are summarized in Table~\ref{table:behavior-expert-ablations}. Optical flow provides additional information on object motions and thus facilitate the prediction of future object locations in normal scenarios. As a result, by comparing the first and the second row, a large improvement in AUC is observed with the inclusion of optical flow features in the input. Additionally, by encoding the perceptual information in the RGB image, the behavior expert is made aware of the environment and is able to further enhance the AD performance. Although the improvement is minor compared to the addition of optical flows, we choose to keep the image pipeline in the behavior expert as the backbone MobileNetV2 has been shown efficient for online tasks~\citep{orsic2019defense}. More importantly, we found that the scaling of anomaly scores with bounding box heights is essential to achieve good AD performance. Without the scaling, the AUC with all three input modalities is even lower than that with only bounding boxes as input but with scaling. Such a phenomenon is due to the fact that STD values of bounding box coordinates may not faithfully reflect the prediction consistency of a model, as detailed in Section~\ref{subsec:bm} and Figure~\ref{fig:height-scaling-eg}.

Figure~\ref{fig:behavior-expert-ablation} further demonstrates the effect of score scaling in two videos with non-ego out-of-control anomalies (OO-N). In both examples, the anomalous car starts swerving in the distance and keeps moving towards the ego car. No matter how far the anomalous object is, the anomaly is present as long as the object can be detected by the camera. Without scaling, however, the score from the behavior expert remains low at the beginning of the anomaly as STD values of the predicted bounding box coordinates are generally small for distant objects. By contrast, the score rectified by bounding box heights can indicate the anomaly more confidently than without scaling in both cases, leading to a clearer separation between the normal and the anomaly. As a result of the ablation study, we adopt the third row in Table~\ref{table:behavior-expert-ablations} with all the inputs and score scaling as the behavior expert in Xen-SIB.

\subsection{Model complexity}
We further conduct a comparative study on model complexity of different methods. Focusing on anomaly detection, we exclude the complexity of other auxiliary tasks (e.g., object detection, optical flow estimation, depth estimation) in each method as these tasks are already performed on self-driving cars~\citep{shi2022csflow,godard2019digging,du2020online}. The results are summarized in Table~\ref{table:model-complexity}. In general, methods that require direct computation on image pixels has a larger number of parameters and FLOPs than those relying on bounding boxes. Compared to the other ensemble method, FOL-ensemble, Xen-SIB has fewer parameters but more FLOPs, mainly due to the scene expert. Notably, the three experts in Xen-SIB are independent of each other by design and can make inference simultaneously. As a result, the inference time of Xen-SIB will be the maximum of that of the three experts. Furthermore, the scene expert can also be parallelized to improve the efficiency bottleneck in Xen-SIB, as FFP (Figure~\ref{fig:ffp}) and STR (Figure~\ref{fig:str}) are independent.

\begin{table}[t]
  \begin{center}
    \caption{Model complexity of different methods with hyperparameters specified in the experiments.}
    \label{table:model-complexity}
    \begin{tabular}{  l  c  c  }
      \toprule
      Method & \# params & FLOPs \\
      \midrule
      Conv-AE & $11.3$M & $114.9$G \\
      AnoPred & $7.7$M & $42.2$G \\
      FOL & $5.2$M & $32.8$M \\
      FOL-Ensemble & $12.9$M & $42.2$G \\
      \midrule
      Scene expert & $8.1$M & $57.9$G \\
      Interaction expert & $0.2$M & $0.5$M \\
      Behavior expert & $3.5$M & $10.8$M \\
      \midrule
      Xen-SIB & $11.8$M & $57.9$G \\
      \bottomrule
    \end{tabular}
  \end{center}
\end{table}

\subsection{Anomaly type classification by unsupervised learning}
Although no class labels on the anomaly type were used during training, Xen-SIB is able to classify videos by taking advantage of the experts, which are specialized for different anomaly patterns. As an extension to bring additional insight to the detection, we introduce binary classification of ego involved and non-ego involved anomalies for each video. The outputs of such a classifier can have practical implications: an ego involved anomaly requires immediate action from the ego car while a non-ego involved anomaly may only need extra attention.

To construct a classifier from Xen-SIB, we make use of the first four states of Kalman filter. Specifically, for a video of length $T$, we take the mean of the highest $10\%$ of the values along the time axis for each state $x_{1:T}^\text{exp}$, denoted as $m^\text{exp}$. Under the assumption that the scene expert mainly captures ego-involved anomalies and the interaction and behavior expert mainly capture non-ego involved anomalies, the decision is made by:
\begin{equation}
\label{eq:classifier-decision}
y
=
\begin{cases}
\text{ego involved,} & m^\text{ffp} + m^\text{str} > m^\text{int} + m^\text{beh}\\
\text{non-ego involved,} & m^\text{ffp} + m^\text{str} <= m^\text{int} + m^\text{beh}
\end{cases}
\end{equation}
where $y$ is the predicted class label.

We evaluate the classifier on all the test videos. The results are summarized in Table~\ref{table:confusion-matrix-classifier}. It has been shown in prior works that traffic anomaly classification is challenging, with all the benchmarked \textit{supervised} methods suffering from low accuracy on DoTA~\citep{yao2022dota}. The problem is made more difficult and seemingly impossible under the setting of unsupervised learning in this paper. Nevertheless, Xen-SIB still performs much better than a random classifier, achieving $67.3\%$ and $57.7\%$ accuracy for ego involved and non-ego involved anomalies, respectively. With a closer observation, we found that the scene expert can detect some of the non-ego involved anomalies when the anomalous object is proximate to the ego car and occupies a large region of the frame. Meanwhile, the behavior expert can also produce high anomaly scores for some of the ego involved anomalies when the object's motion in the image plane becomes unpredictable due to the sudden change of the ego car's dynamics. As a result of the above two phenomena, the assumption behind the decision~(\ref{eq:classifier-decision}) is not always true for all driving videos, leading to a misclassification of the anomaly type.

\begin{table}[t]
  \begin{center}
    \caption{Confusion matrix over all the test videos.}
    \label{table:confusion-matrix-classifier}
    \resizebox{\linewidth}{!}{%
    \begin{tabular}{ c  l  c  c }
      \multicolumn{2}{c}{} & \multicolumn{2}{c}{\bfseries Predicted} \\
      \multicolumn{2}{c}{} & ego involved & non-ego involved \\
      \cmidrule{2-4}
      \bfseries Actual & ego involved & $67.3$ & $32.7$ \\
      & non-ego involved & $42.3$ & $57.7$ \\
      \cmidrule{2-4}
    \end{tabular}}
  \end{center}
\end{table}

%% file: Sections/07-Discussion.tex
\section{Discussion and limitations}
The results presented in Section~\ref{sec:experiments} show that compared to baseline methods, Xen provides a more effective solution that can enable an autonomous car to detect on-road anomalies in diverse driving scenarios using a single monocular camera. Despite the advantages, our work also encompasses several limitations.

Object detection plays an important role in Xen, as the interaction and behavior expert work under the assumption that anomalous objects can be reliably detected for trajectory reconstruction and prediction, respectively. However, the perception capability of monocular cameras is largely limited when the visibility is poor, such as at night and under inclement weather conditions. To alleviate the issue, camera-LiDAR fusion has been proposed and shown more effective than unimodal approaches in computer vision tasks~\citep{cui2021deep,chen2017multi,sindagi2019mvx}. With an additional sensor modality, object detection and thus anomaly detection can be made more robust in different environments. Furthermore, as noted in Section~\ref{subsec:bm}, perspective projection onto the image plane distorts the motion characteristics of objects (e.g., a proximate object appears to move faster than a distant object even though the two objects have the same speed in reality), which challenges efficient modeling of normal motion patterns. 3D object detection enabled by point clouds from LiDAR has the potential to resolve the issue by projecting bounding boxes to bird's eye view (BEV) and thus eliminating the negative effect of perspective projection on learning object motions.

Another common failure case of Xen results from large scene motions in normal scenarios, e.g., when the ego car executes an aggressive lane change or moves fast in complex urban areas. Frame prediction becomes difficult in such cases due to large motions of the ego car, and the resulting increase of score is indistinguishable from that caused by an anomaly. It has been shown recently in video prediction literature that camera poses are helpful in rendering high-quality images~\citep{ak2021robust}. As a result, given that additional onboard vehicle state is available, ego motions can be exploited to create a more robust scene expert. Another similar issue that can cause false positives in Xen is discussed in supplemental materials.

Anomaly detection is an active research topic both in robotics and computer vision. At a more general level, we hope that the analysis in this work, especially those in Section~\ref{sec:overview}, provides insights on a unified framework for anomaly detection in related areas. More specifically, an anomaly detector can be designed based on Figure~\ref{fig:anomaly-patterns} with necessary modifications for different applications. For example, for AD on field robots which operate in autonomous farms without human labors, only the edge between the ego agent and the environment needs to be monitored as the robot often performs a task individually; for AD with surveillance cameras, the two edges with one of the ends being the ego agent can be ignored as the surveillance camera is fixed and will never participate in an anomaly; and for AD on mobile robots that navigate through human crowds, the whole graph needs to be considered if non-ego involved anomalies also affect robot decisions. With the high-level framework determined, each expert can then be designed specifically for each type of edge based on the characteristics of different anomalies. Validating the generalization capability of Xen in other application domains is left as future work.

Another possible direction is to evaluate the efficacy of more complex architectures, such as foundation models, for anomaly detection. Large visual language models (LVLMs) have been shown powerful in a variety of application areas, including image captioning, content generation, and conversational AI~\citep{jiang2024effectiveness}. In the domain of autonomous driving, LVLMs have also been explored for tasks of visual question-answering~\citep{xu2024drivegpt4}, trajectory prediction~\citep{wu2023language}, path planning~\citep{mao2023language}, and decision-making and control~\citep{wen2023road}. These recent research advancements suggest that incorporating foundation models into on-road anomaly detection is a promising direction.

Although powerful, LVLMs are currently limited in efficiency due to billions of parameters~\citep{brohan2022rt,brohan2023rt,padalkar2023open}. Furthermore, proprietary models, such as GPT-4V, must be queried over the cloud, further increasing inference time~\citep{achiam2023gpt}. To ensure both accuracy and efficiency of on-road anomaly detection with limited onboard resources, a combination of LVLMs and lightweight models is necessary. One integration method is to retrieve intermediate embeddings of images through the LVLM, which can then be provided as an additional context to lightweight anomaly detectors for inference. The embeddings from the LVLM can be updated periodically for efficiency. Such a method, however, requires access to the hidden states of the LVLM, which most proprietary models do not allow. Alternatively, LVLMs can be used as an additional anomaly detection expert, which can then be incorporated into Xen through the Kalman filter. While the three original experts update the system states of Kalman filter at a high frequency, the LVLM can be queried at a low frequency and updates the system states asynchronously in a similar manner. With such an approach, we are able to benefit from both the efficiency of lightweight models and the accuracy and generalization ability of LVLMs.

%% file: Sections/08-Conclusion.tex
\section{Conclusion}
We present an ensemble method for anomaly detection in autonomous driving using a single monocular camera. We provide a holistic analysis on common anomaly patterns in driving scenarios, which is then used to design experts or modules tuned to detect different types of anomalies. The scene expert aims to capture frame-level abnormal events, which are often accompanied with ego involved anomalies; the interaction expert models normal relative motions between two road participants and raises an alarm whenever anomalous interactions emerge in non-ego involved interactive anomalies; and the behavior expert detects non-ego involved individual anomalies by monitoring inconsistent predictions of future locations for each object. A Kalman filter is developed for expert ensemble, in which the observations are normalized expert scores and the final anomaly score is modeled as one of the states. Our experimental results, with a novel evaluation protocol to reflect realistic model performance, show that the proposed method outperforms various baselines with higher AUC and F1-score in diverse driving scenarios in DoTA. Furthermore, we demonstrate that the multi-expert design can also enable the classification of an on-road anomaly using unsupervised learning. With several possible directions to explore in the future, we hope that this work provides insights on approaching anomaly detection in related areas through anomaly pattern analysis and can serve as a competitive solution to 2D on-road anomaly detection for autonomous driving.